\documentclass{article}

\usepackage[preprint]{neurips_2025}

\usepackage[utf8]{inputenc} 
\usepackage[T1]{fontenc}    
\usepackage{hyperref}       
\usepackage{url}            
\usepackage{booktabs}       
\usepackage{amsfonts}       
\usepackage{algorithm}

\usepackage{algpseudocode}

\usepackage{amsmath}        
\usepackage{nicefrac}       
\usepackage{microtype}      
\usepackage{xcolor}         
\usepackage{natbib} 

\usepackage{multirow}  
\usepackage{makecell}  
\usepackage{colortbl}
\usepackage{booktabs}  
\usepackage{colortbl}  
\usepackage{multirow}  
\usepackage{graphicx}

\usepackage[bb=dsserif]{mathalpha}
\usepackage{bm}

\usepackage[capitalize,noabbrev,nameinlink]{cleveref}
\definecolor{cornflowerblue}{rgb}{0.39, 0.58, 0.93}
\hypersetup{
    colorlinks=true,
    linkcolor=cornflowerblue,
    filecolor=magenta,      
    urlcolor=teal,
    citecolor=cornflowerblue,
    pdftitle={GPTailor: Large Language Model Pruning Through Layer Cutting and Stitching},
    }

\makeatletter
\renewcommand{\alglinenumber}[1]{}
\makeatother
\algdef{S}{NoLine}{}{}
\algtext*{NoLine}

\title{GPTailor: Large Language Model Pruning Through Layer Cutting and Stitching
}

\author{
 Guinan Su\textsuperscript{1},
 \And
 Li Shen\textsuperscript{5},
 \And
 Lu Yin\textsuperscript{6},
 \And
 Shiwei Liu\textsuperscript{7},
 \And
  Yanwu Yang\textsuperscript{4},
  \And
  Jonas Geiping\textsuperscript{1,2,3}
  \AND
  \textsuperscript{1}\normalfont Max Planck Institute for Intelligent Systems,
  \textsuperscript{2}\normalfont ELLIS Institute Tübingen \\
  \textsuperscript{3}Tübingen AI Center,
  \textsuperscript{4}University of Tübingen,
  \textsuperscript{5}Sun Yat-sen University
  \\
  \textsuperscript{6}University of Surrey,
  \textsuperscript{7}University of Oxford
}

\begin{document}

\maketitle

\begin{abstract}
Large language models (LLMs) have shown remarkable capabilities in language understanding and generation. However, such impressive capability typically comes with a substantial model size, which presents significant challenges in deployment and inference. While structured pruning of model parameters offers a promising way to reduce computational costs at deployment time, current methods primarily focus on single model pruning. In this work, we develop a novel strategy to compress models by strategically combining or merging layers from finetuned model variants, which preserves the original model's abilities by aggregating capabilities accentuated in different finetunes. We pose the optimal tailoring of these LLMs as a zero-order optimization problem, adopting a search space that supports three different operations: (1) Layer removal, (2) Layer selection from different candidate models, and (3) Layer merging. Our experiments demonstrate that this approach leads to competitive model pruning, for example, for the Llama2-13B model families, our compressed models maintain approximately 97.3\% of the original performance while removing $\sim25\%$ of parameters, significantly outperforming previous state-of-the-art methods. The code is available at.\footnote{\url{https://github.com/Guinan-Su/auto-merge-llm}}

\end{abstract}

\renewcommand{\thefootnote}{}
\footnote{Correspondence to: shenli6@mail.sysu.edu.cn}
\renewcommand{\thefootnote}{\arabic{footnote}} 

\section{Introduction}
The unique strengths of modern Large Language Models (LLMs) in language understanding, generation, and reasoning \cite{touvron2023llama, achiam2023gpt, chiang2023vicuna} are inextricably linked to their immense size.
Research in this field has generally followed a trajectory of scaling model parameters and data to enhance performance, guided by two fundamental principles: scaling laws, which establish that performance improves predictably with increased parameters \cite{kaplan2020scaling, hoffmann2022training, wei2022emergent}, and over-parameterization theory, which demonstrates that models with excess parameters achieve better optimization and generalization \cite{allen2019learning, allen2019convergence, li2020train}. These principles have led researchers to develop billion-parameter architectures delivering unprecedented performance across diverse language tasks.

Despite these impressive capabilities, deploying LLMs presents significant challenges due to their substantial computational demands. Various post-training techniques have been proposed to address the issues faced when deploying models to consumer GPUs or local devices, or when reducing costs, including model pruning \cite{frantar2023sparsegpt, dettmers2023spqr, xia2023sheared, kim2024shortened, ma2023llm}, knowledge distillation into smaller models \cite{chen2022disco, hsieh2023distilling, shridhar2023distilling, tunstall2023zephyr}, and quantization of weights \cite{yao2022zeroquant, gholami2022survey, dettmers2023qlora}. While quantization reduces parameter precision but requires specific hardware support, and knowledge distillation necessitates costly retraining of smaller models, structured pruning offers a more flexible and hardware-agnostic approach by eliminating redundant parameters to decrease computation costs.

Existing pruning methods typically focus on pruning individual models through manually designing metrics that assess the importance of specific structures or layers based on hidden state changes or gradient information \cite{kim2024shortened, men2024shortgpt, ma2023llm}. However, these approaches inevitably cause performance degradation and require additional post-training with full parameters to recover performance.

\begin{figure*}
    \centering
    \vspace{-0.4cm}
    \includegraphics[width=1.0\textwidth]{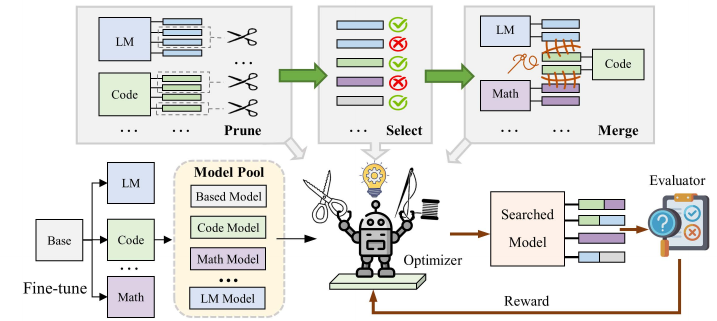}
    \caption{Our Approach: \textbf{Model Pruning through Cutting and Stitching}. We achieve competitive model pruning performance by running a zero-order search that tailors layers based on a shared pool of finetuned variants of the original model, selecting and stitching layers if necessary. The model finetunes accentuate task-specific skills, allowing us to merge key components into a smaller model, maintaining, for example, 97\% of capabilities of Llama-13B, even after a 25\% reduction in layers.}
    \vspace{-0.4cm}
    \label{fig:fig1}
\end{figure*}

%
To address these limitations, we take a radically different perspective and re-formulate structured pruning as the problem of \textit{pruning not individual models, but a family of task-specific finetuned versions of a given model}. These finetuned variants are surprisingly helpful for model pruning, as each variant accentuates a particular task, such as coding, math, or language understanding. Further, the variants are close enough that model merging can be employed to re-combine layers from multiple variants, if needed \cite{wortsman2022model}. These observations lead us to our main question:
\textbf{Can we develop better compressed models by strategically combining or merging layers from different models?} Motivated by this question, we propose a novel structured pruning method based on zero-order optimization that supports three different operations to combine layers from different models into a smaller, more efficient model: \textbf{(1) Layer removal, (2) Layer selection from related candidate models, (3) Layer merging}.

For the optimization, we define multiple objective functions that capture different aspects of model performance across different tasks to better preserve the original model's capabilities and run a fully data-driven zero-order optimization, instead of relying on expert-made heuristics for pruning. We employ SMAC~\cite{JMLR:v23:21-0888}, which strategically allocates computational resources by evaluating configurations at different calibration data sizes, thereby reducing computational costs while boosting the efficiency of finding superior solutions. We rigorously validate our method's effectiveness by evaluating it on Llama-7B and Llama-13B with four state-of-the-art structural pruning methods across comprehensive benchmarks. Our experimental results demonstrate that our approach maintains excellent performance while outperforming existing pruning methods.

In summary, the main contributions of this paper are:
\begin{itemize}
    \item We propose a novel structured pruning method that formulates pruning as a zero-order optimization problem over a pool of candidate models, enabling automated discovery of efficient models that leverage capabilities from multiple models.
    \item We find that this approach allows for a cost-effective model pruning stage that is effective without the need for post-training to heal the pruned model.
    \item We validate our method's effectiveness through extensive experiments, comparing against modern LLM pruning methods on 14 benchmark tasks. 
\end{itemize}
\looseness -1 Our method maximally preserves the capabilities of the dense model: 92.2\% for the 7B model and 97.3\% for the 13B model. significantly outperforming previous state-of-the-art methods. 




\section{Related Work}

\textbf{Compression of Language Models.}
Large language models \citep{touvron2023llama,achiam2023gpt,chiang2023vicuna} necessitate efficient compression methods to reduce parameters and latency. These methods include structural pruning \citep{frantar2023sparsegpt,dettmers2023spqr,xia2023sheared,kim2024shortened,ma2023llm}, knowledge distillation \cite{chen2022disco, hsieh2023distilling,tunstall2023zephyr}, and quantization \cite{yao2022zeroquant,gholami2022survey,dettmers2023qlora}.
Our work focuses on structural pruning, which removes sub-components from neural networks for hardware-friendly compression - instead of pruning through sparsification, which requires significant effort to materialize gains on standard hardware. Recent methods focus on pruning via expert-designed criteria, include LLMPruner \cite{ma2023llm}, which removes non-critical structures using gradient information; SliceGPT \cite{ashkboos2024slicegpt}, which replaces weight matrices with smaller ones to reduce dimensions; LaCo \cite{yang2024laco}, 
which prunes by collapsing the weights of later layers into earlier layers based on activation similarity, 
and ShortGPT \cite{men2024shortgpt}, which uses Block Influence (BI) to measure layer importance based on hidden state changes.
Unlike these metric-based methods targeting individual models, our approach employs zero-order \textit{search}, namely hyperparameter optimization to combine pruning and merging across model families. While LaCo also uses layer merging, it focuses only on merging similar layers for a single model, whereas we focus on strategically combining or merging layers from different models, which we find to noticeably improve upon within-model merging. Additionally, our approach differs from the weight-sharing NAS-based pruning method \cite{klein2024structural}, requiring costly super-network training; our approach directly optimizes across fine-tuned models, strategically combining layers from diverse variants rather than searching within a single model.

\textbf{Model Merging.}
Model merging enhances capabilities without additional training data or computation. The field evolved from simple weighted parameter averaging \cite{utans1996weight} that often yielded suboptimal results to advanced techniques like Task Arithmetic \cite{ilharco2022editing} which computes differences between model parameters and SLERP \cite{white2016sampling} which performs interpolation along spherical paths. Later approaches leveraged neural network sparsity, with TIES-Merging \cite{yadav2024ties} selecting parameters based on magnitude while addressing sign conflicts, and DARE \cite{yu2024language} combining sparsification with parameter rescaling. Recent advances include Evolutionary model merging \cite{akiba2024evolutionary} optimizing coefficients through evolutionary search, and multi-fidelity approach \cite{su2025fine} that enables fine-grained exploration while reducing costs. Our work also builds upon a multi-fidelity optimization framework to allow for an efficient search for compressed models.

\section{Methods}

In this section, we provide a detailed explanation of our approach. Unlike conventional model compression pipelines, we formulate pruning as a zero-order optimization problem over the layers and merging hyperparameters of a set of candidate models. 
We begin in \cref{sec:problem_setup} by outlining our problem formulation and defining the optimization pipeline for pruning with three key components: a search space, a target objective, and an optimizer. \Cref{sec:search_space} follows with a description of the search spaces. In \cref{sec:objective}, we introduce our designed target objective function. Finally, In \cref{sec:optimizer}, we describe our choice of optimization strategy, which efficiently navigates the defined search space to identify optimal pruning configurations. An overview of the pipeline is provided in \cref{fig:fig1}.

\subsection{Problem Setup}\label{sec:problem_setup}
Given a pre-trained base model $M_{\text{base}}$ and a set of candidate models $\mathcal{M} = \{M_1, M_2, ..., M_K\}$ fine-tuned from the same base model, our goal is to find an optimal pruned model that maximizes performance while adhering to a target sparsity constraint. Let $s$ denote the target sparsity factor, where $s\in [0,1]$ indicates the fraction of parameters to be pruned.
The pruned model is constructed through a combination of layers from candidate models, employing operations such as layer-wise merge, layer selection, and layer removal. These combinations and operations are determined by a set of hyperparameters $\omega \in \Omega$, with $\Omega$ representing the search space of all possible hyperparameter configurations. Each configuration $\omega$ defines a specific way to combine the layers from candidate models to form a pruned model $M_\omega$.
The performance of the pruned model can be evaluated using a function $f(M_\omega)$, which measures the model's effectiveness on specific datasets and tasks. This leads to our optimization problem:
\begin{equation}
\omega^* = \arg\min_{\omega \in \Omega} f(M_\omega) \quad \text{subject to} \quad \text{S}(M_\omega) \leq s
\end{equation}
where $\text{S}(\cdot)$ calculates the fraction of pruned parameters in the model compared to the base model, and $\omega^*$ represents the optimal hyperparameter configuration that yields the performing pruned model.

\subsection{Search Space Design}\label{sec:search_space}

The search space $\Omega$ encompasses all possible pruning configurations that can be applied to construct our pruned model. We formulate this space based on structural layer-wise pruning operations. We aim to support three operations: (1) Layer removal, (2) layer selection, and (3) Layer merging. We designed our search space as follows:

Given a base model with $l$ layers and $K$ candidate models fine-tuned from this base model, we design the search space through a binary vector $\mathbf{r} = [r_1, r_2, \ldots, r_l]$ where $r_i \in \{0,1\}$ indicates whether the $i$-th layer is retained ($r_i = 0$) or removed ($r_i = 1$), satisfying $\sum_{i=1}^{l} r_i = \lceil l \cdot s \rceil$ to achieve our target sparsity $s$. For each retained layer position $i$, we define a selection vector $\mathbf{c}_i = [c_{i,1}, c_{i,2}, \ldots, c_{i,K}]$ where $c_{i,j} \in \{0,1\}$ indicates whether the layer from the $j$-th candidate model is selected. If $\sum_{j=1}^{K} c_{i,j} = 0$, we retain the layer from the base model instead. When multiple candidate models contribute to a layer position (i.e., $\sum_{j=1}^{K} c_{i,j} > 1$), we specify a merge method $m_i \in \{1, 2, \ldots, Z\}$ from $Z$ available merging techniques. Each merge method $m_i$ is associated with a set of hyperparameters $\mathbf{h}_i = [h_{i,1}, h_{i,2}, \ldots, h_{i,P_i}]$, where $P_i$ is the number of hyperparameters for the specific merge method. These hyperparameters govern the precise mechanism of layer combination, such as interpolation weights or mask ratio parameters. Therefore, a complete configuration $\omega \in \Omega$ is represented as $\omega = \{\mathbf{r}, \{\mathbf{c}_i | r_i = 0\}, \{m_i | r_i = 0 \text{ and } \sum_{j=1}^{K} c_{i,j} > 1\}, \{\mathbf{h}_i | r_i = 0 \text{ and } \sum_{j=1}^{K} c_{i,j} > 1\}\}$. 
The total cardinality of the search space can be calculated as:
$|\Omega| = \binom{l}{\lceil l \cdot s \rceil} \times \prod_{i: r_i = 0} 2^K \times \prod_{i: r_i = 0, \sum_{j=1}^{K} c_{i,j} > 1} Z \times \prod_{i: r_i = 0, \sum_{j=1}^{K} c_{i,j} > 1} |\mathbf{h}_i|$. which enables a wide exploration of pruning strategies while maintaining the target sparsity constraint.



\subsection{Target Objective Function}\label{sec:objective}

To evaluate the quality of a pruned model, we define a multi-objective function that measures the model's effectiveness across tasks. Specifically, we measure performance on calibration datasets $\mathcal{D}_{\text{calibration}}$, quantifying metrics such as accuracy for classification tasks or perplexity for language modeling tasks. This provides a direct assessment of how well the pruned model preserves the capabilities of the original model. We define a multi-task objective function that captures different aspects of model performance across a range of tasks to produce a comprehensive pruned model. Let $\mathcal{T} = \{T_1, T_2, \ldots, T_m\}$ be a set of $m$ tasks. For a pruned model $M_\omega$ with configuration $\omega$, we employ Pareto Efficient Global Optimization (ParEGO)\citep{knowles2006parego} to identify Pareto-optimal solutions across different objectives. Specifically, the ParEGO algorithm transforms multi-objective optimization problems into a series of single-objective problems through scalarization methods:


\begin{equation}
f_{multi}(M_\omega, \lambda) = \max_{i=1,\ldots,m} \{\lambda_i \cdot f_i(M_\omega)\} + \alpha \sum_{i=1}^{m} \lambda_i \cdot f_i(M_\omega)
\end{equation}

where $f_i(M_\omega)$ is the $i$-th objective function, $\lambda_i$ is the corresponding weight satisfying $\sum_{i=1}^{m} \lambda_i = 1$ and $\lambda_i \geq 0$, and $\alpha$ is a small positive constant (typically set to 0.05). The Chebyshev norm component $\max_{i=1,\ldots,m} \{\lambda_i \cdot f_i(M_\omega)\}$ ensures that all non-dominated solutions on the non-convex Pareto front can be identified, while the term $\alpha \sum_{i=1}^{m} \lambda_i \cdot f_i(M_\omega)$ enhances the algorithm's stability. The final output of our optimizer is a Pareto front of pruning configurations, where each configuration represents a different trade-off between performance on various tasks. In our experiments, we randomly selected three configurations from this Pareto front and report their results.

\subsection{Search Optimizer}\label{sec:optimizer}
To efficiently navigate the search space and find optimal pruning configurations, we employ SMAC \citep{JMLR:v23:21-0888}, which strategically allocates computational resources by evaluating configurations at different fidelity levels. we use calibration dataset size as fidelity type, represented by budgets b where $b_{\text{min}} \leq b \leq b_{\text{max}}$. Each budget value corresponds to a specific portion of the calibration data used for evaluation - smaller budgets (lower fidelity) use fewer samples for faster but less precise evaluations, while larger budgets (higher fidelity) use more samples for slower but more accurate assessments. We use Random Forest \cite{breiman2001random} as a surrogate model to sample new configurations. Given configuration space $\Omega$, minimum budget $b_{\min}$, maximum budget $b_{\max}$, reduction factor $\eta$ and the maximum trials $T_{\max}$, the whole process is described in \Cref{alg:successive_halving}.





\begin{algorithm}
\caption{The optimization process of pruning.}
\label{alg:successive_halving}
\begin{algorithmic}[1]
\Require Configuration space $\Omega$, minimum budget $b_{\min}$, maximum budget $b_{\max}$, reduction factor $\eta$, maximum trials $T_{\max}$
\Ensure Optimized configuration $\omega^*$
\State $s_{\max} = \lfloor \log_{\eta} \frac{b_{\max}}{b_{\min}} \rfloor$, $D \gets \emptyset$, $T \gets 0$ \Comment{Initialization}

\For{$s \in \{s_{\max}, s_{\max}-1, \ldots, 0\}$ and $T < T_{\max}$}
    \State $n \gets \lceil \frac{(s_{\max}+1)}{(s+1)} \cdot \eta^s \rceil$, $r \gets b_{\min} \cdot \eta^s$ \Comment{Config count \& budget}
    \State $\mathcal{C} \gets$ Sample Configurations($n$, $D$, $\Omega$) \Comment{Sample configurations}
    
    \For{$i \in \{0, 1, \ldots, s\}$ and $T < T_{\max}$}
        \State $n_i \gets \lfloor n \cdot \eta^{-i} \rfloor$, $r_i \gets r \cdot \eta^i$ \Comment{Stage parameters}
        
        \For{each $w \in \mathcal{C}$ and $T < T_{\max}$}
            \State Evaluate $y_{w} \gets f_{\text{multi}}(M_w, \lambda, r_i)$, $D \gets D \cup \{(w, r_i, y_{w})\}$, $T \gets T + 1$
        \EndFor
        
        \State Sort $\mathcal{C}$ by performance, keep the top $\lfloor n_i / \eta \rfloor$ configurations in $\mathcal{C}$
    \EndFor
\EndFor
\State \Return the best-performing configuration $\omega^*$ evaluated at highest budget
\end{algorithmic}
\end{algorithm}

This efficient optimization strategy enables us to handle the search space defined in \Cref{sec:search_space}, identifying high-performing pruned models that satisfy our multi-objective function from \cref{sec:objective}, with significantly reduced computational cost compared to exhaustive search approaches.


\section{Experiments}

\begin{table}[htbp]
\centering
\scriptsize
\caption{Comparison of pruning methods on multiple natural language benchmarks. "Single" refers to the best performance achieved when pruning a single model directly, while "Merge" refers to the best performance achieved through either "pruning-then-merging" or "merging-then-pruning". For 7b model:  Llama-2-7B-Chat (LM), MAmmoTH-7B (Math), Llama-2-Coder-7B (Code), and Llama-2-7B (Base), for 13b model:  WizardLM-13B (LM), WizardMath-13B (Math), llama-2-13b-
code-alpaca (Code), and Llama-2-13B (Base). The cells highlighted in blue show three selected Pareto-optimal solutions of our method.}
\definecolor{lightgray}{rgb}{0.9,0.9,0.9}
\definecolor{lightblue}{rgb}{0.8,0.85,1.0}
\setlength{\tabcolsep}{0.8pt} 
\renewcommand{\arraystretch}{1.1} 
\begin{tabular}{ccc|ccc|ccc|cccc|cccc|c|c}
\hline\hline
\rowcolor{gray!16}
\textbf{LLM} & \textbf{Pruner} & \textbf{Type} & \multicolumn{3}{c|}{\textbf{Reasoning}} & \multicolumn{3}{c|}{\textbf{Language}} & \multicolumn{4}{c|}{\textbf{Knowledge}} & \multicolumn{4}{c|}{\textbf{Understanding}} & \multirow{1}{*}{\textbf{Avg}} & \multirow{1}{*}{\textbf{Avg*}} \\
\rowcolor{gray!16}
 & \textbf{(ratio)} &  & \textbf{CNLI} & \textbf{HeSw} & \textbf{PIQA} & \textbf{CHID} & \textbf{WSC$_P$} & \textbf{WSC$_G$} & \textbf{CSQA} & \textbf{BoolQ} & \textbf{MMLU} & \textbf{CMLU} & \textbf{Race$_H$} & \textbf{Race$_M$} & \textbf{XSum} & \textbf{C3} & &\\
\hline
\multirow{12}{*}{\makecell{Llama\\-7B}} & \multirow{4}{*}{\makecell{Dense\\(0.0\%)}} & Base & 32.98  & 71.34 & 78.18 & 41.56 & 37.50 & 38.46 & 55.04 & 70.70 & 46.67 & 31.88 & 35.53 & 33.36 & 19.55 & 43.84 & 45.47 & 42.30 \\
& & Math & 32.99 & 68.60 & 75.79 & 39.71 & 39.42 & 36.54 & 50.78 & 69.36 & 43.04 & 32.16 & 30.36 & 36.42 & 20.88 & 43.45 & 44.25 & 41.70 \\
& & LM & 31.30 & 71.28 & 75.95 & 36.11 & 63.46 & 59.62 & 64.29 & 74.77 & 48.30 & 33.93 & 52.52 & 55.22 & 22.45 & 47.56 & 52.63 & 47.24 \\
& & Code & 32.99 & 70.27 & 78.62 & 41.61 & 36.54 & 41.35 & 57.41 & 71.04 & 46.22 & 32.20 & 41.25 & 39.69 & 18.79 & 46.25 & 46.73 & 43.79 \\
\cline{2-19}


& \multirow{2}{*}{\makecell{LLMPru\\(25.3\%)}} & Single & 32.99 & \textbf{59.57} & \textbf{73.34} & \textbf{30.32} & 46.15 & 0.00 & 20.15 & 57.28 & 23.21 & 25.16 & 21.56 & 21.52 & \textbf{15.19} & 31.07 & 32.68 & 32.74 \\
& & Merge & 34.71 & \textbf{60.57} & \textbf{73.50} & 26.62 & 40.38 & 5.77 & 19.90 & 52.14 & 24.01 & 25.30 & 23.07 & 22.98 & \textbf{15.51} & 32.49 & 32.64 & 32.60 \\
\cline{2-19}


& \multirow{2}{*}{\makecell{SliceGPT\\(26.3\%)}} & Single & 31.89 & 41.55 & 58.81 & 18.43 & 39.42 & 4.81 & 19.49 & 40.09 & 25.38 & 25.02 & 25.59 & 26.88 & 8.78 & 39.56 & 28.98 &28.64\\
& & Merge & 32.85 & 37.61 & 57.56 & 17.33 & 53.85 & 2.88 & 19.41 & 42.66 & 25.22 & 24.68 & 25.21 & 24.72 & 12.78 & 40.22 & 29.78 & 28.67 \\
\cline{2-19}

& \multirow{2}{*}{\makecell{LaCo\\(27.1\%)}} & Single & 32.97 & 55.24 & 69.53 & \textbf{31.47} & 36.54 & 34.62 & 22.11 & 67.22 & 29.08 & 26.16 & 28.53 & 28.27 & 14.68 & \textbf{43.51} & 37.14 & 36.45 \\
& & Merge & 31.89 &  56.26 &  \textbf{71.22} &  \textbf{27.32} &  39.42 &  22.12 &  23.42 &  72.66  &  29.30 &   26.00 &  25.19 &  26.81 &  \textbf{16.11} &  \textbf{43.62} &  36.52 & 36.21\\
\cline{2-19}

& \multirow{2}{*}{\makecell{ShortGPT\\(27.1\%)}} & Single & 33.09 & 57.42 & 66.54 & 21.53 & 56.73 & \textbf{48.08} & 52.50 & 67.34 & 43.68 & 28.31 & 32.53 & 31.69 & 12.40 & 39.45 & 42.24 & 35.97 \\
& & Merge & 34.10 & 54.18 & 64.42 & 16.83 & 61.54 & 36.54 & 55.61 & \textbf{73.21} & 36.84 & 25.61 & 42.94 & 45.89 & 10.12 & 35.73 & 42.40 & 37.62 \\
\cline{2-19}

& \multirow{3}{*}{\makecell{Ours\\(27.1\%)}} &  & \cellcolor{lightblue}\textbf{35.46} & \cellcolor{lightblue}54.43 & \cellcolor{lightblue}67.74 & \cellcolor{lightblue}23.63 & \cellcolor{lightblue}\textbf{63.46} & \cellcolor{lightblue}\textbf{43.27} &\cellcolor{lightblue} \textbf{62.90} & \cellcolor{lightblue}\textbf{75.08} & \cellcolor{lightblue}\textbf{48.75} & \cellcolor{lightblue}\textbf{33.86} & \cellcolor{lightblue}\textbf{55.35} & \cellcolor{lightblue}\textbf{58.64} & \cellcolor{lightblue}12.99 & \cellcolor{lightblue}\textbf{44.16} & \cellcolor{lightblue}48.55 & \cellcolor{lightblue} 43.73\\
& &  & \cellcolor{lightblue}\textbf{34.94} & \cellcolor{lightblue}\textbf{58.14} & \cellcolor{lightblue}69.48 & \cellcolor{lightblue}21.53 & \cellcolor{lightblue}\textbf{63.46} & \cellcolor{lightblue}41.35 & \cellcolor{lightblue}\textbf{62.74} & \cellcolor{lightblue}66.24 & \cellcolor{lightblue}\textbf{47.39} &\cellcolor{lightblue}\textbf{34.11} & \cellcolor{lightblue}\textbf{49.17} & \cellcolor{lightblue}\textbf{50.56} & \cellcolor{lightblue}3.46 & \cellcolor{lightblue}{41.53} & \cellcolor{lightblue}46.01 & \cellcolor{lightblue} 39.96\\
& &  & \cellcolor{lightblue}\textbf{34.95} & \cellcolor{lightblue}54.92 & \cellcolor{lightblue}67.08 & \cellcolor{lightblue}24.48 & \cellcolor{lightblue}\textbf{63.46} & \cellcolor{lightblue}\textbf{46.15} &\cellcolor{lightblue} \textbf{62.00} & \cellcolor{lightblue}\textbf{75.90} & \cellcolor{lightblue}\textbf{48.73} & \cellcolor{lightblue}\textbf{34.13} & \cellcolor{lightblue}\textbf{54.03} & \cellcolor{lightblue}\textbf{57.45} & \cellcolor{lightblue}13.20 & \cellcolor{lightblue}{43.01} & \cellcolor{lightblue}48.54 & \cellcolor{lightblue}43.56 \\
\cline{2-19}

\hline
\multirow{12}{*}{\makecell{Llama\\-13B}} & \multirow{4}{*}{\makecell{Dense\\(0.0\%)}} & Base & 32.99 & 74.77 & 79.71 & 47.35 & 50.96 & 63.46 & 67.24 & 71.38 & 55.84 & 38.74 & 57.98 & 60.17 & 23.47 & 47.51 & 55.11 &  50.48\\
& & LM & 35.36 & 70.41 & 78.73 & 36.21 & 57.69 & 60.58 & 65.03 & 73.70 & 53.48 & 30.85 & 66.12 & 71.66 & 22.44 & 52.00 & 55.30 & 50.97 \\
& & Math & 32.99 & 68.78 & 77.26 & 44.36 & 36.54 & 19.23 & 60.36 & 78.44 & 54.21 & 38.12 & 47.74 & 48.82 & 19.51 & 44.66 & 47.93 &  47.05\\
& & Code & 32.99 & 74.82 & 80.14 & 47.30 & 51.92 & 63.46 & 68.88 & 72.72 & 55.92 & 39.26 & 58.03 & 63.72 & 24.45 & 48.38 & 55.86 & 51.30 \\
\cline{2-19}

& \multirow{2}{*}{\makecell{LLMPru\\(21.2\%)}} & Single & \textbf{33.49} & 60.28 & \textbf{75.57} & 23.68 & 39.42 & 0.00 & 19.00 & 63.24 & 23.27 & 25.23 & 22.36 & 21.45 & \textbf{17.13} & 32.00 & 32.58 & 33.21 \\
& & Merge & \textbf{33.86} & 64.11 & \textbf{73.50} & 22.18 & \textbf{60.58} & 0.00 & 21.46 & 61.96 & 23.84 & 25.62 & 22.16 & 21.59 & 14.98 & 32.11 & 34.14 & 33.17 \\
\cline{2-19}

& \multirow{2}{*}{\makecell{SliceGPT\\(23.6\%)}} & Single & \textbf{33.19} & 42.44 & 59.90 & 18.03 & 54.81 & 19.23 & 32.51 & 41.22 & 33.09 & 25.75 & 29.45 & 29.87 & 9.99 & 37.75 & 33.37 & 29.74 \\
& & Merge & 30.98 & 46.83 & 62.57 & 19.33 & 51.92 & 49.04 & 37.76 & 38.38 & 33.55 & 25.22 & 23.53 & 23.05 & 9.95 & 39.67 & 35.13 & 28.55\\
\cline{2-19}

& \multirow{2}{*}{\makecell{LaCo\\(24.6\%)}} & Single & 32.33 & 60.18 & 70.57 & \textbf{32.67} & 34.62 & 34.62 & 52.58 & 62.66 & 36.26 & 25.80 & 60.38 & 62.53 & 8.79 & \textbf{49.21} & 44.51 & 43.84 \\
& & Merge & \textbf{33.49} & 62.50 & 74.37 &\textbf{35.26} & \textbf{63.46} & \textbf{63.46} & 18.84 & 64.65 & 41.83 & 24.87 & 26.10 & 25.97 & 15.93 & 39.51 & 42.16 & 34.71\\
\cline{2-19}

& \multirow{2}{*}{\makecell{ShortGPT\\(24.6\%)}} & Single & 32.95 & 62.64 & \textbf{73.50} & 28.22 & 36.54 & 50.96 & 65.44 & 67.71 & 53.50 & 30.73 & \textbf{65.52} & \textbf{71.38} & \textbf{19.12} & \textbf{48.60} & 50.49 & 47.43 \\
& & Merge & 31.07 & 63.24 & 68.61 & 27.17 & 49.04 & 43.27 & 65.68 & \textbf{78.01} & 51.26 & 36.88 & 57.38 & 62.67 & \textbf{16.94} & 44.05 & 49.66 & 46.38 \\
\cline{2-19}

& \multirow{3}{*}{\makecell{Ours\\(24.6\%)}} &  & \cellcolor{lightblue}32.99 & \cellcolor{lightblue}\textbf{66.81} & \cellcolor{lightblue}\textbf{75.03} & \cellcolor{lightblue}29.07 & \cellcolor{lightblue}54.81 & \cellcolor{lightblue}\textbf{62.50} & \cellcolor{lightblue}\textbf{69.37} & \cellcolor{lightblue}\textbf{74.28} &\cellcolor{lightblue} \textbf{55.90} & \cellcolor{lightblue}\textbf{39.71} & \cellcolor{lightblue}\textbf{65.52} & \cellcolor{lightblue}\textbf{71.03} & \cellcolor{lightblue}16.80 & \cellcolor{lightblue}46.74 & \cellcolor{lightblue}54.33 & \cellcolor{lightblue}49.22 \\
& &  & \cellcolor{lightblue}31.80 & \cellcolor{lightblue}\textbf{68.63} & \cellcolor{lightblue}72.52 & \cellcolor{lightblue}30.97 & \cellcolor{lightblue}\textbf{60.58} & \cellcolor{lightblue}\textbf{55.77} & \cellcolor{lightblue}\textbf{67.49} & \cellcolor{lightblue}\textbf{73.70} & \cellcolor{lightblue}\textbf{54.61} & \cellcolor{lightblue}\textbf{39.29} & \cellcolor{lightblue}61.92 & \cellcolor{lightblue}\textbf{70.13} & \cellcolor{lightblue}16.19 & \cellcolor{lightblue}48.11 & \cellcolor{lightblue}53.69 & \cellcolor{lightblue} 48.97 \\
& &  & \cellcolor{lightblue}29.05 & \cellcolor{lightblue}\textbf{69.76} & \cellcolor{lightblue}72.74 & \cellcolor{lightblue}\textbf{34.22} & \cellcolor{lightblue}\textbf{58.65} & \cellcolor{lightblue}54.81 & \cellcolor{lightblue}\textbf{68.06} & \cellcolor{lightblue}69.82 & \cellcolor{lightblue}\textbf{53.99} & \cellcolor{lightblue}\textbf{38.36} & \cellcolor{lightblue}\textbf{62.32} & \cellcolor{lightblue}66.71 & \cellcolor{lightblue}16.60 & \cellcolor{lightblue}\textbf{51.01} & \cellcolor{lightblue}53.29 & \cellcolor{lightblue}48.65 \\
\cline{2-19}

\hline\hline
\end{tabular}
\label{tab:mainres}
\vspace{-0.3cm}
\end{table}
\subsection{Experimental Settings}

\textbf{Benchmarks.} 
To evaluate the pruned model’s capabilities, we utilized the OpenCompass evaluation framework \cite{2023opencompass}. Specifically, we conduct evaluations in five aspects: Reasoning, Language, Knowledge, Examination and Understanding. Reasoning: CMNLI (CNLI)\cite{xu2020clue}, HellaSwag (HeSw)\cite{zellers2019hellaswag}, PIQA \cite{bisk2020piqa}. Language: CHID \cite{zheng2019chid}, WSC \cite{levesque2012winograd}. Knowledge: CommonSenseQA (CSQA) \cite{talmor2018commonsenseqa}, BoolQ \cite{clark2019boolq}. Examination: MMLU \cite{hendrycks2020measuring}, CMMLU (CMLU) \cite{li2023cmmlu}. Understanding: Race-High/Middle (H/M) \cite{lai2017race}, XSum \cite{narayan2018don}, C3 \cite{sun2020investigating}. For CHID and XSum, we use generative evaluation. For the WSC dataset, we use cloze log-likelihood (WSCP) and generative (WSCG) evaluation. The remaining benchmarks are evaluated using cloze log-likelihood. See detailed benchmark information in \Cref{sec:b}.

\textbf{Baselines.}
To evaluate the effectiveness of our method, we compared with four state-of-the-art structured pruning methods: LLM-Pruner (LLMPru) \cite{ma2023llm}, SliceGPT \cite{ashkboos2024slicegpt}, LaCo \cite{yang2024laco}, and ShortGPT \cite{men2024shortgpt}.  In our experiments, we set the pruning ratios of our method to be equivalent to ShortGPT and LaCo, and slightly higher than others to ensure fair comparison. Furthermore, as our method is based on multiple candidate models, we check three comprehensive comparison scenarios to guarantee fairness: (1) Applying each baseline pruning method individually to all candidate models and picking the strongest one, (2)  First pruning each candidate model using the baseline method and then merging them, and (3) First merging the candidate models and then applying pruning.
For model merging across all baseline experiments, we employ the same task-arithmetic merging~\cite{ilharco2022editing} technique used in our search space, with merging factors within the range $[0.5, 1.0]$~\cite{ilharco2022editing}.

\textbf{Model Selection.}
To assess the effectiveness of the proposed method, we search for pruned versions of the popular Llama2-7B and Llama2-13B \cite{touvron2023llama}. For 7B models, we use Llama-2-7B \cite{touvron2023llama} as our base model, with three candidate models: Llama-2-7B-Chat \cite{touvron2023llama} (LM), MAmmoTH-7B \cite{yue2023mammoth} (Math), and Llama-2-Coder-7B \cite{manuel_romero_2023} (Code). For 13B models, we use Llama-2-13B \cite{touvron2023llama} as the base model, with WizardLM-13B \cite{xu2023wizardlm} (LM), WizardMath-13B \cite{luo2023wizardmath}(Math), and Llama-2-13B-Code-Alpaca \cite{codealpaca} (Code) as candidate models.
For the 7B models, we set the sparsity ratio to 9/32, removing approximately 28\% of the layers. For the 13B models, we set the sparsity ratio to 10/40, removing approximately 25\% of the layers. These two ratios are matching the best settings from prior work in ShortGPT and LaCo, while being slightly higher than other baseline methods, allowing for fair comparisons. For layer merging, we implement task-arithmetic \cite{ilharco2022editing} merging with a configurable merging factor that controls the magnitude of task-specific adaptations.


\textbf{Calibration Data.}
For our calibration dataset, we selected multiple-choice datasets to ensure the model's generalization ability across different capabilities. Specifically, we sampled from diverse datasets: 1000 examples from the PIQA \cite{bisk2020piqa} training set, 500 examples from the WSC \cite{levesque2012winograd} training set, 1000 examples from the CSQA \cite{talmor2018commonsenseqa} training set, and 1000 examples from the MMLU \cite{hendrycks2020measuring} validation set (which is distinct from the MMLU test set). This diverse collection allows us to calibrate our model across a broad spectrum of linguistic and reasoning capabilities.

\textbf{Objective and Optimizer.}
Our implementation builds upon SMAC \cite{JMLR:v23:21-0888} for optimization. We allocate 500 search trials for both 13B and 7B experiments. To improve optimization efficiency, we use models with randomly removed middle layers as starting points, since models are relatively robust to changes in these intermediate layers \cite{su2025fine}. We set the minimum budget $b_{\text{min}}$ as 100, maximum budget $b_{\text{max}}$ as the 1000, and reduction factor $\eta$ as 3. This resulted in budgets of \{100, 300, 1000\} for  PIQA, CSQA, and MMLU. For the WSC, we set budgets to \{100, 200, 500\}

\subsection{Main Results}
To validate the efficiency of our method, we compared it with the four baselines: LLM-Pruner (LLMPru) \cite{ma2023llm}, SliceGPT \cite{ashkboos2024slicegpt}, LaCo \cite{yang2024laco}, and ShortGPT \cite{men2024shortgpt}. We reproduce the results from these methods and evaluate on OpenCompass~\cite{2023opencompass}. As mentioned, to validate that our proposed approach of \textit{"pruning while merging"}  is optimal, we also re-run each pruning method on (1) pruning each candidate model individually and picking the best, (2) "pruning-then-merging": First pruning each candidate model using the baseline method and then merging them, and (3) "merging-then-pruning": First merging the candidate models and then applying pruning. 


\Cref{tab:mainres} reports the best single model pruning and best merge results of all baselines, with full results in \Cref{sec:d}. We selected three Pareto-optimal solutions from our results. Our approach achieves the best results across multiple benchmarks compared to all tested LLM pruning methods. In terms of overall performance, our method maximally preserves the capabilities of the dense model: 92.2\% (48.55/52.63) for the 7B model and 97.3\% (54.33/55.86) for the 13B model. To ensure our results were not biased by our calibration data, we also calculate an avg* excluding the four benchmarks from which training data was selected for calibration (MMLU, CSQA, WSC, PIQA). As shown in the avg* column, our method still outperformed all baselines, further validating our approach. Notably, our method achieved comparable or even better results than dense models on many benchmarks. We attribute these gains to: 1) Pruning might mitigate "overthinking" \citep{kaya2019shallow} effects, evident in benchmarks like CNLI and WSC where other baseline pruning methods also improved performance, and 2) our merging strategy is effectively compensating for information loss from pruning.

\begin{figure*}
    \centering
    \vspace{-0.8cm}
    \includegraphics[width=1.0\textwidth]{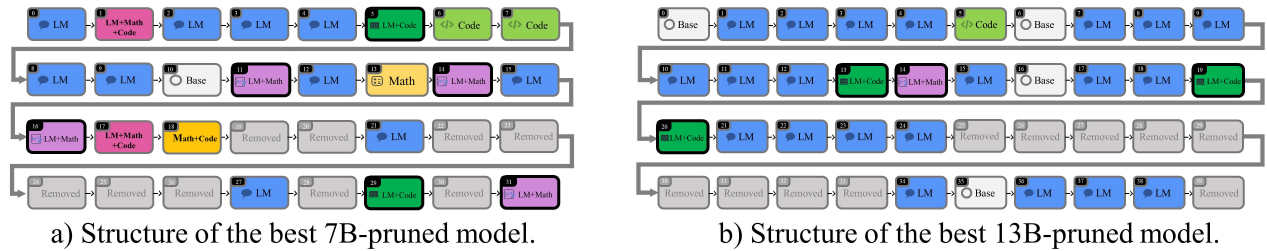}
    \caption{(a) Structure of our best-performing 7B-pruned model. The model integrates layers from multiple candidates: Llama-2-7B-Chat (LM), MAmmoTH-7B (Math), Llama-2-Coder-7B (Code), and Llama-2-7B (Base). The pruning ratio is 9/32, removing 9 layers out of 32 total layers. (b) Structure of our best-performing 13B-pruned model. The model integrates layers from multiple candidates:  WizardLM-13B (LM), WizardMath-13B (Math), llama-2-13b-code-alpaca (Code), and Llama-2-13B (Base). The pruning ratio is 10/40, removing 10 layers out of 40 total layers.}
    \label{fig:arch7b}
    \vspace{-0.4cm}
\end{figure*}

\Cref{fig:arch7b} illustrates our best-performing 7B-pruned model and best-performing 13B-pruned models's structure (See \Cref{tab:arch7} and \Cref{tab:arch13} for architectural details). We observe that both models tend to remove middle-to-later layers, with the 13B model removing layers from layer 25 and the 7B model from layer 19. This suggests information redundancy in these layers, aligning with findings that later layers exhibit high similarity and redundancy~\cite{men2024shortgpt,gromov2024unreasonable}. The 13B model shows a simpler structure dominated by a single LM model with concentrated layer removal, while the 7B model shows a more complex structure utilizing mixed and specialized models with scattered layer removal. This suggests that as model size decreases, more diverse mixing strategies may be needed to maintain performance.  This architectural difference, coupled with the superior preservation rate of the 13B model compared to the 7B model, demonstrates that robustness (redundancy) scales with model size.

\subsection{Efficiency Analysis}
\begin{figure}[htbp]
    \centering
  
\begin{minipage}{0.48\textwidth}
    \centering
     \vspace{-0.6cm}
    \begin{table}[H]
        \centering
        \scriptsize
        \setlength{\tabcolsep}{4pt}
        \begin{tabular}{c|cc|ll}
            \toprule
            \textbf{Search Budget} & \textbf{Percentage} & \textbf{Trials} & \textbf{Dataset} & \textbf{Size} \\
            \midrule
            \multirow{4}{*}{Low} & \multirow{4}{*}{41.4} & \multirow{4}{*}{207} & PIQA & 100 \\
            & & & WSC & 100 \\
            & & & CSQA & 100 \\
            & & & MMLU & 100 \\
            \midrule
            \multirow{4}{*}{Medium} & \multirow{4}{*}{36.6} & \multirow{4}{*}{183} & PIQA & 300 \\
            & & & WSC & 200 \\
            & & & CSQA & 300 \\
            & & & MMLU & 300 \\
            \midrule
            \multirow{4}{*}{High} & \multirow{4}{*}{22.0} & \multirow{4}{*}{110} & PIQA & 1000 \\
            & & & WSC & 500 \\
            & & & CSQA & 1000 \\
            & & & MMLU & 1000 \\
            \bottomrule
        \end{tabular}
    \vspace{0.1cm}
          \caption{Budget allocation to search trials for pruning. $41\%$ of trials require only the smallest budget size, significantly increasing efficiency.}
        \label{tab:trial_dist}
    \end{table}
\end{minipage}
\hfill 
\begin{minipage}{0.48\textwidth}
    \centering
    \vspace{-0.8cm}
    \includegraphics[width=\textwidth]{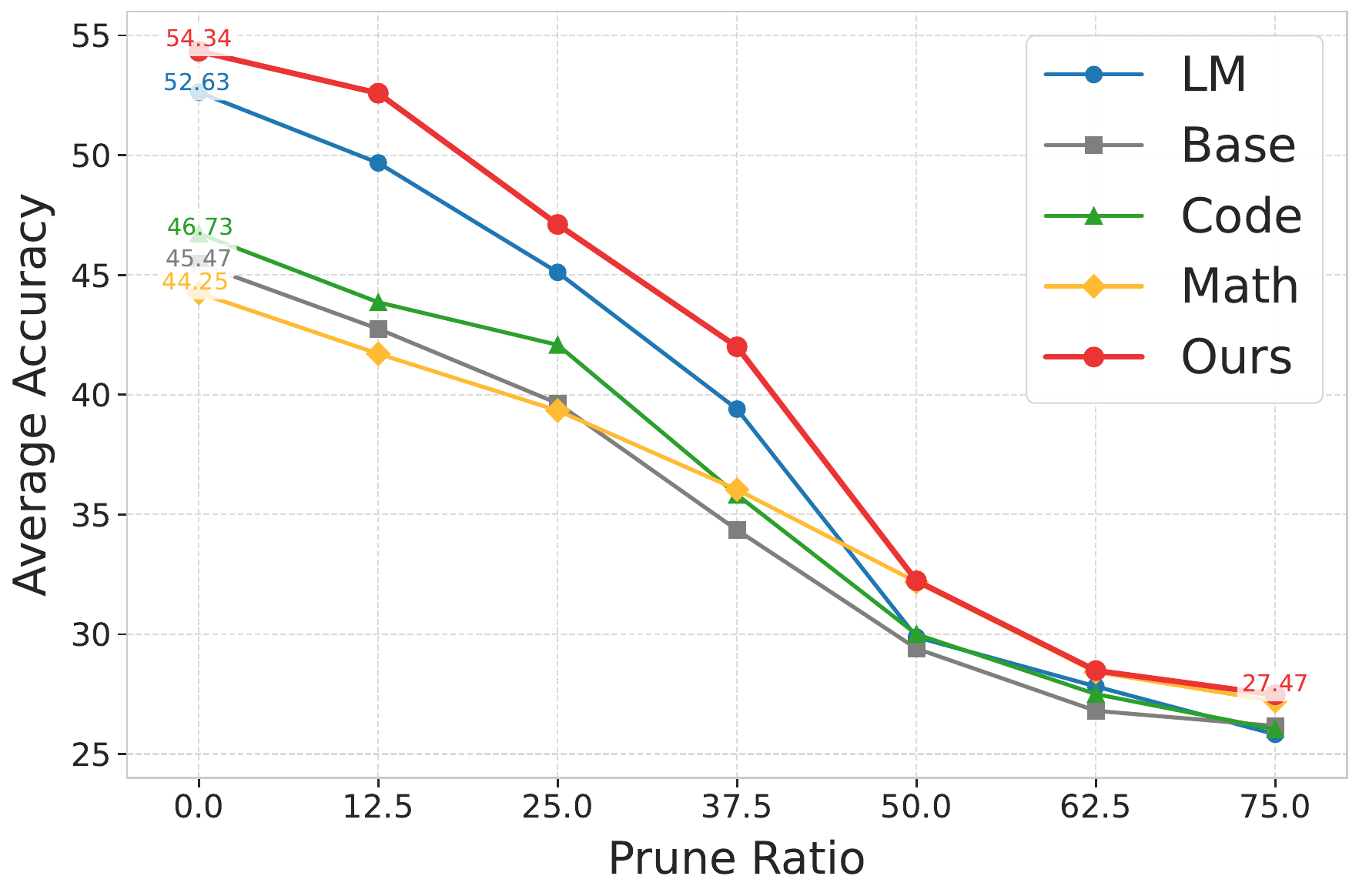}
    \vspace{-0.6cm}
     \caption{Performance Comparison Across Different Pruning Ratios.}
     \label{fig:fig2}
\end{minipage}
\vspace{-0.5cm}
\end{figure}

Our optimizer dynamically adjusts the budget allocation during the search process,
where the budget is defined as the calibration dataset size
used for search. As the allocation of search trials directly determines the overall search duration. Here, we analyze the budget distribution
during the search process, as shown in \cref{tab:trial_dist}.
Our analysis reveals that only 22\% of the search trials utilize the full budget, while
over 41.4\% of the evaluations were conducted with the minimum budget, which is 5-10 times smaller.  This efficient allocation enables our pruning to significantly increase the chance of discovering superior configurations under the same computational budget.

\subsection{Which Parts of the Search Space are Critical ?}

To determine where the benefits of our approach come from, we designed ablation experiments to evaluate the contribution of different components in our search space. As our framework supports: (1) selectively choosing layers from different candidate models, (2) layer merging, and (3) layer removal, we conducted several experiments to isolate the impact of each component. \Cref{tab:model_comparison} summarizes the performance comparison across various benchmarks (full results available in the \Cref{tab:model_comparisonfull})

\textbf{Layer Removal Only (LR-only).} We restricted the search space to only allow layer removal operations on a single 7B model. We ran experiments on all 7B models and report the best performer. This ablation shows a performance drop from our full approach (48.55 → 44.83 on average), confirming that merely pruning layers from a single model is insufficient for optimal performance. The performance degradation is particularly notable on language tasks (WSC$_P$: 63.46 → 49.04) and understanding benchmarks (Race$_H$: 55.35 → 42.51, Race$_M$: 58.64 → 43.04). It is worth highlighting that even our layer-removal-only for single model still outperforms the strongest baseline method, ShortGPT (44.83 vs. 42.24). This demonstrates that our approach can enhance performance even in this simplified setting.

\textbf{Layer Selection and Removal (LS+LR).} In this setting, we enabled both layer selection from different candidate models and layer removal operations but disabled the layer merging functionality. The results show even greater performance degradation (48.55 → 43.20 average) compared to the layer-removal-only setting. We observe a dramatic drop on WSC$_G$ (43.27 → 26.92), indicating that merging operations play a critical role for certain grammatical reasoning tasks. The superior performance of LR-only (44.83) compared to LS+LR (43.20) demonstrates that simply combining layers from different models without proper integration through merging is suboptimal.

\definecolor{lightgray}{rgb}{0.9,0.9,0.9}
\definecolor{lightblue}{rgb}{0.8,0.85,1.0}
\setlength{\tabcolsep}{1.3pt} 
\renewcommand{\arraystretch}{1.2} 
\begin{table}[htbp]
\caption{Comparison of different searching settings across various benchmarks. Settings: LR-only: Layer-remove only, LS+LR: Layer-selection + layer-remove, FL-merge: Folding Layers Merging, Single-obj: Single-objective, PPL-obj: PPL as search objective.}
\centering
\scriptsize
\begin{tabular}{l ccc ccc cccc cccc c}
\toprule
\rowcolor{gray!16}
\textbf{Setting} & \multicolumn{3}{c}{\textbf{Reasoning}} & \multicolumn{3}{c}{\textbf{Language}} & \multicolumn{4}{c}{\textbf{Knowledge}} & \multicolumn{4}{c}{\textbf{Understanding}} & \textbf{Avg} \\
\cmidrule(lr){2-4} \cmidrule(lr){5-7} \cmidrule(lr){8-11} \cmidrule(lr){12-15}
\rowcolor{gray!16}
 & \textbf{CNLI} & \textbf{HeSw} & \textbf{PIQA} & \textbf{CHID} & \textbf{WSC$_P$} & \textbf{WSC$_G$} & \textbf{CSQA} & \textbf{BoolQ} & \textbf{MMLU} & \textbf{CMLU} & \textbf{Race$_H$} & \textbf{Race$_M$} & \textbf{XSum} & \textbf{C3} & \\
\midrule
Ours & 35.46& 54.43&67.74 &23.63 &63.46 & 43.27&62.90 & 75.08& 48.75&33.86 & 55.35&58.64&12.99 &44.16 &48.55 \\
LR-only &34.96 &53.80&66.70 &18.58&49.04 &58.65 &60.61&68.87&47.85&33.54&42.51&43.04&8.05 &41.42 &44.83\\
LS+LR&32.92&55.84&65.07 &17.98 &63.46&26.92& 58.97&51.22& 48.97&34.61&48.68&49.44& 8.33&42.41& 43.20\\

FL-merge&32.99 &52.90&63.66&19.28 &46.15&62.50&60.52 &75.20&48.30&34.33 &50.77 &55.29& 6.39&39.40&46.26 \\

Single-obj &32.15 &56.02 &67.46&19.08&39.42 &48.08&62.33& 74.43&47.40& 34.14&50.94&52.86&12.35&41.97&45.62\\
PPL-obj &33.39&23.89 &52.07& 14.84& 45.19&7.69 &19.33&39.51 & 24.25&24.69 &22.81&21.17&0.06&26.36&25.38\\
\bottomrule
\end{tabular}
\label{tab:model_comparison}
\vspace{-0.5cm}
\end{table}

\subsection{Additional Analysis}

\subsubsection{Varying the Pruning Ratio}
To evaluate the benefits of evolving from "pruning single model" to "pruning from model variants" under varying pruning ratios, we compared our approach with each candidate model using the layer-remove configuration, as it achieves the strongest performance among single-model pruning methods, even surpassing the best-performing baseline, ShortGPT. \Cref{fig:fig2} visualizes the average accuracy among benchmark performances at different pruning ratios, with detailed results in \Cref{tab:prune_ratio}. The accuracy of all models decreases as the pruning ratio increases. Our model achieves the best performance at almost all pruning ratios, especially in the low pruning ratio range of 0 - 37.5\%. When pruning reaches 50\%, the performance gap narrows across all models as they all experience performance collapse, including ours. We believe this is due to excessive parameter removal, after which effective model function cannot be maintained without additional post-training.

\subsubsection{Pruning through Layer Folding}
LaCo \cite{yang2024laco} is another pruning approach based on merging, but it differs from our method by 
merging only the later layers of a single model into adjacent earlier layers, based on activation similarity heuristics.
To validate the effectiveness and potential of this type of within-model merge operation, we use our hyperparameter optimization framework with a specially designed search space consisting of: (1) A binary selection vector $\mathbf{s} = [s_1, s_2, \ldots, s_k]$ indicating which layers to remove, and (2) An importance weight vector $\mathbf{w} = [w_1, w_2, \ldots, w_k]$ representing each layer's importance value.

Retained layer $L_i'$ performs a depth-wise linear combination with itself and adjacent removed layers:
$$L_i' = \beta_i \cdot L_i + \sum_{j \in \mathcal{N}(i)} \beta_j \cdot L_j \cdot \mathbb{1}_{s_j = 1}$$
where $\mathcal{N}(i)$ represents adjacent layers to $L_i$, $\mathbb{1}_{s_j = 1}$ indicates layer $j$ is removed, and $\beta_j$ are normalized weights derived from $\mathbf{w}$ such that $\beta_i + \sum_{j \in \mathcal{N}(i)} \beta_j \cdot \mathbb{1}(s_j = 1) = 1$. This ensures retained layers incorporate information from nearby removed layers, preserving network functionality. This configuration parametrizes the options proposed in LACO and achieves better overall performance (46.26) compared to previous ablations but still falls short of our full approach. The performance gap is most pronounced on XSum (12.99 → 6.39) and PIQA (67.74 → 63.66), highlighting the importance of our optimized merging strategy for generative and reasoning tasks.

\vspace{-0.1cm}
\subsubsection{Different Calibration Datasets and Metrics}
\vspace{-0.1cm}
In our method, we use multiple-choice datasets as calibration data with accuracy as the metric in a multi-objective optimization approach. This results in a pruned model with broad capabilities. To further analyze this design choice, we conducted the following experiments:

\textbf{Single Objective (Single-obj).}
We used the MMLU validation dataset for calibration with accuracy as the optimization objective. We evaluate the resulting pruned models across our benchmark suite. As shown in \cref{tab:model_comparison}, while these models still perform adequately (45.62 average), the single-objective optimization led to a noticeable decline from our full approach (48.55 → 45.62). Importantly, the single-objective models demonstrated stronger performance on MMLU-related tasks but showed performance degradation on certain other tasks due to their narrow optimization focus. This confirms our hypothesis that broad, multi-objective optimization is necessary to preserve the broad functionality of modern LLMs, rather than overfitting to a single task domain.

\textbf{Perplexity Objective (PPL-obj).}
We also experiment with perplexity (PPL) on WikiText \citep{merity2016pointer} as a search metric, taking 1500 examples as our calibration dataset. The performance of resulting pruned models across benchmarks in \cref{tab:model_comparison} demonstrates a dramatic performance gap compared to all other configurations, with an average score of only 25.38 across benchmarks. Even when compared to the single-objective MMLU optimization (which uses a similarly sized dataset), the PPL-optimized models showed considerably weaker performance across most tasks. The catastrophic degradation on tasks like XSum (12.99 → 0.06) and reasoning benchmarks like HeSw (54.43 → 23.89) underscores the limitations of perplexity-guided optimization. This evidence reveals that while perplexity is a common metric for language model evaluation, it fails to serve as an effective signal for preserving model capabilities during pruning, particularly for tasks requiring reasoning and knowledge application rather than just fluent text generation.

\subsubsection{Extending to Llama-3}
We further extend our validation to Meta's Llama 3 8B model\cite{grattafiori2024llama}. Llama 3 training on 15 trillion tokens---7$\times$ more data than Llama 2, incorporates architectural improvements including universal Grouped Query Attention (GQA), an optimized 128K vocabulary tokenizer and longer context window. These modifications further boost the performance on reasoning, code generation, and multilingual tasks. Despite similar model size, Llama-3 8B achieves better performance compared to Llama-2 7B\cite{touvron2023llama}, which may carry different semantic densities that provide new challenges for maintaining model performance under compression. Validating on this next-generation model is crucial for establishing the practical applicability of our method in rapidly evolving LLM landscapes.

we use Meta-Llama-3-8B\cite{grattafiori2024llama} as our base model, with three candidate models: Meta-Llama-3-8B-Instruct (LM)\cite{grattafiori2024llama}, Code-Llama-3-8B (Code)\cite{codellama3-8b}, and MathCoder2-Llama-3-8B (Math)\cite{wang2024mathcoder,lu2024mathcoder2bettermathreasoning}. We target removing 9 of the 32 layers and use the same experimental settings as Llama2-7B.

We compare our method with the best-performing baseline, ShortGPT. As results shown in \cref{tab:8bres}(full results available in the \cref{tab:8bfullres}), our method retains 84.55\%(53.17/63.61) of the original performance after removing 9 layers, outperforming ShortGPT's  62.79\%(39.94/63.61) retention under same compression ratios. while both show lower retention than our Llama2-7B results (92.2\%) with similar model size. This decline indicates Llama3's reduced compressibility,  We attribute this decline to (1) higher parameter utilization efficiency and (2) denser knowledge distribution from large-scale training, which eliminates layer redundancy. Despite this challenge, our method consistently outperforms the baseline, validating its effectiveness across model generations.

\begin{table}[H]
\centering
\scriptsize
\caption{Comparison of pruning methods on multiple natural language benchmarks. "Single" refers to the best performance achieved when pruning a single model directly, while "Merge" refers to the best performance achieved through either "pruning-then-merging" or "merging-then-pruning". For 8b model:   Meta-Llama-3-8B-Instruct (LM), MathCoder2-Llama-3-8B (Math), Code-Llama-3-8B (Code), and Meta-Llama-3-8B (Base). The cells highlighted in blue show three selected Pareto-optimal solutions of our method.}
\setlength{\tabcolsep}{0.8pt} 
\renewcommand{\arraystretch}{1.3} 
\definecolor{lightgray}{rgb}{0.9,0.9,0.9}
\definecolor{lightblue}{rgb}{0.8,0.85,1.0}

\begin{tabular}{ccc|>{\columncolor{white}}c>{\columncolor{white}}c>{\columncolor{white}}c|>{\columncolor{white}}c>{\columncolor{white}}c>{\columncolor{white}}c|>{\columncolor{white}}c>{\columncolor{white}}c>{\columncolor{white}}c>{\columncolor{white}}c|>{\columncolor{white}}c>{\columncolor{white}}c>{\columncolor{white}}c>{\columncolor{white}}c|>{}c}
\hline\hline
LLM & Pruner & Type & \multicolumn{3}{c|}{Reasoning} & \multicolumn{3}{c|}{Language} & \multicolumn{4}{c|}{Knowledge} & \multicolumn{4}{c|}{Understanding} & \multirow{2}{*}{Avg} \\
 & ratio/layer &  & CMNLI & HeSw & PIQA & CHID & WSC$_P$ & WSC$_G$ & CSQA & BoolQ & MMLU & CMLU & Race$_H$ & Race$_M$ & XSum & C3 & \\
\hline
\multirow{8}{*}{\makecell{Llama3\\-8B}} & \multirow{4}{*}{Dense} & \cellcolor{lightgray}Base & \cellcolor{lightgray}32.98  & \cellcolor{lightgray}74.67 & \cellcolor{lightgray}80.96 & \cellcolor{lightgray}73.78 & \cellcolor{lightgray}56.73 & \cellcolor{lightgray}36.54 & \cellcolor{lightgray}73.79 & \cellcolor{lightgray}69.97 & \cellcolor{lightgray}64.74 & \cellcolor{lightgray}50.79 & \cellcolor{lightgray}63.21 & \cellcolor{lightgray}70.54 & \cellcolor{lightgray}3.28 & \cellcolor{lightgray}55.18 & \cellcolor{lightgray}57.65\\
& & \cellcolor{lightgray}LM & \cellcolor{lightgray}33.00 & \cellcolor{lightgray}71.08 & \cellcolor{lightgray}80.69 & \cellcolor{lightgray}65.53 & \cellcolor{lightgray}55.77 & \cellcolor{lightgray}69.23 & \cellcolor{lightgray}76.66 & \cellcolor{lightgray}78.87 & \cellcolor{lightgray}65.97 & \cellcolor{lightgray}53.64 & \cellcolor{lightgray}76.44 & \cellcolor{lightgray}81.75 & \cellcolor{lightgray}17.97 & \cellcolor{lightgray}63.95 & \cellcolor{lightgray}63.61\\
& & \cellcolor{lightgray}Math & \cellcolor{lightgray}32.99 & \cellcolor{lightgray}71.66 & \cellcolor{lightgray}77.97 & \cellcolor{lightgray}57.09 & \cellcolor{lightgray}37.50 & \cellcolor{lightgray}58.65 & \cellcolor{lightgray}68.22 & \cellcolor{lightgray}69.08 & \cellcolor{lightgray}62.08 & \cellcolor{lightgray}45.85 & \cellcolor{lightgray}64.75 & \cellcolor{lightgray}69.08 & \cellcolor{lightgray}8.68 & \cellcolor{lightgray}53.86 & \cellcolor{lightgray}55.53\\
& & \cellcolor{lightgray}Code & \cellcolor{lightgray}32.98 & \cellcolor{lightgray}65.56 & \cellcolor{lightgray}74.70 & \cellcolor{lightgray}78.42 & \cellcolor{lightgray}61.54 & \cellcolor{lightgray}61.54 & \cellcolor{lightgray}63.47 & \cellcolor{lightgray}78.35 & \cellcolor{lightgray}48.03 & \cellcolor{lightgray}34.55 & \cellcolor{lightgray}52.40 & \cellcolor{lightgray}58.43 & \cellcolor{lightgray}19.36 & \cellcolor{lightgray}46.41 & \cellcolor{lightgray}55.41\\
\cline{2-18}

& \multirow{2}{*}{\makecell{ShortGPT}} & Single & 32.83 & 45.06 & 65.78 & 23.38 & 41.35 & 53.85 & 39.56 & 63.73 & 32.37 & 28.69 & 40.14 & 45.19 & \textbf{3.68} & 43.51 & 39.94 \\

& & Merge & 32.95 & 48.58 & 64.96 & 18.43 & 36.54 & 35.58 & 42.83 & \textbf{67.22} & 33.05 & 28.71 & 30.16 & 32.45 & \textbf{3.66} & 44.27 & 37.10\\
\cline{2-18}

& \multirow{3}{*}{\makecell{Ours}} &  & \cellcolor{lightblue}\textbf{33.46} & \cellcolor{lightblue}54.81 & \cellcolor{lightblue}\textbf{69.53} & \cellcolor{lightblue}\textbf{32.27} & \cellcolor{lightblue}41.35 & \cellcolor{lightblue}58.65 & \cellcolor{lightblue}72.81 & \cellcolor{lightblue}\textbf{65.29} & \cellcolor{lightblue}\textbf{63.36} & \cellcolor{lightblue}\textbf{50.14} & \cellcolor{lightblue}71.41 & \cellcolor{lightblue}75.97 & \cellcolor{lightblue}3.22 & \cellcolor{lightblue}\textbf{47.12} & \cellcolor{lightblue}52.81\\
& &  & \cellcolor{lightblue}33.07 & \cellcolor{lightblue}\textbf{55.15} & \cellcolor{lightblue}69.37 & \cellcolor{lightblue}31.77 & \cellcolor{lightblue}\textbf{46.15} & \cellcolor{lightblue}\textbf{64.42} & \cellcolor{lightblue}\textbf{73.55} & \cellcolor{lightblue}64.43 & \cellcolor{lightblue}62.35 & \cellcolor{lightblue}49.06 & \cellcolor{lightblue}\textbf{74.36} & \cellcolor{lightblue}\textbf{78.27} & \cellcolor{lightblue}2.95 & \cellcolor{lightblue}46.03 & \cellcolor{lightblue}53.64\\
& &  & \cellcolor{lightblue}\textbf{33.42} & \cellcolor{lightblue}\textbf{54.83} & \cellcolor{lightblue}\textbf{69.75} & \cellcolor{lightblue}\textbf{34.02} & \cellcolor{lightblue}\textbf{47.12} & \cellcolor{lightblue}\textbf{62.50} & \cellcolor{lightblue}\textbf{73.79} & \cellcolor{lightblue}64.34 & \cellcolor{lightblue}\textbf{63.13} & \cellcolor{lightblue}\textbf{50.04} & \cellcolor{lightblue}\textbf{72.81} & \cellcolor{lightblue}\textbf{77.65} & \cellcolor{lightblue}3.00 & \cellcolor{lightblue}\textbf{46.52} & \cellcolor{lightblue}53.78\\
\hline\hline
\end{tabular}
\label{tab:8bres}
\end{table}

\section{Conclusion}
In this work, we presented a novel LLM compression approach that strategically combines layers from fine-tuned model variants instead of pruning single models. By formulating this as a zero-order optimization problem with a newly designed search space that supports layer removal, selection, and merging, our method effectively preserves model capabilities while reducing size. Experiments on Llama2-7B and Llama2-13B demonstrated that our compressed models retain 92.2\% and 97.3\% of original performance, respectively,  despite removing $\sim25\%$ of parameters, outperforming previous state-of-the-art methods without requiring expensive post-training. Overall, our work demonstrates that cutting and stitching layers from multiple fine-tuned variants of a model is a more effective approach to LLM compression than traditional single-model pruning. While the search complexity increases with the number of candidate models, this computational aspect represents an opportunity for future optimization techniques to further enhance efficiency.

\section*{Acknowledgements}
The authors thank Alexander Panfilov and Niccolò Ajroldi for their insightful comments and valuable feedback on this work. JG acknowledges the support of the Hector II foundation. GS acknowledges the support of the International Max Planck Research School for Intelligent Systems (IMPRS-IS).

\bibliographystyle{unsrt} 
\bibliography{refer}

\appendix

\section{Baseline} \label{sec:a}

To ensure fair comparison, we applied various baseline pruning methods including LLM-Pruner(LLMPru) \cite{ma2023llm}, SliceGPT \cite{ashkboos2024slicegpt}, LaCo \cite{yang2024laco} and ShortGPT \cite{men2024shortgpt}: 

\textbf{LLM-Pruner}  adopts structural pruning that selectively removes non-critical coupled structures based on gradient information, maximally preserving the majority of the LLM’s functionality. It applies post-training to the pruned model, for fair comparison, we do not apply post training to it.

\textbf{SliceGPT} is a post-training sparsification scheme that replaces each weight matrix with a smaller matrix, reducing the embedding dimension of the network. Specifically, they applied PCA to the hidden representation from shallow to deep layers, and incorporated the dimension reduction matrix into existing network parameters. 

\textbf{LaCo} is a pruning method for large language models based on reducing layers. LaCo gradually merges similar layers from deep to shallow and sets a threshold to avoid merging too many layers.

\textbf{ShortGPT} introduced the Block Influence (BI) metric, which uses the similarity between layer's input and output to measure the importance of each layer.

\section{Evaluation Benchmarks}  \label{sec:b}

\textbf{CMNLI (Chinese Multi-Genre Natural Language Inference) (CNLI)} consists of two parts: XNLI and MNLI. It contains text from various domains, including fiction, telephone conversations, travel, and government sources.  XNLI is a cross-lingual extension of the MultiNLI corpus, professionally translated into multiple languages, including Chinese, providing a robust framework for assessing language understanding across linguistic boundaries. Models must determine whether pairs of sentences exhibit entailment, contradiction, or neutrality.

\textbf{HellaSwag (HeSw)} tests commonsense reasoning about physical situations. The dataset uses a "Goldilocks" zone of complexity where examples are obviously nonsensical to humans but challenging for state-of-the-art models. Despite being trivial for humans (>95\% accuracy), even advanced models struggled with this benchmark upon its release, making it effective for measuring progress in commonsense inference.

\textbf{PIQA (Physical Interaction Question Answering)} Developed by Bisk et al. (2020), this multi-choice question and answer dataset focuses on everyday scenarios, exploring models' understanding of real-world physical laws through daily situations.

\textbf{CHID (Chinese IDiom)} is an idiom cloze test focusing on the representation and selection of Chinese idioms, requiring cultural and linguistic knowledge specific to Chinese.

\textbf{WSC (Winograd Schema Challenge)} serves as a prominent benchmark for evaluating machine understanding through pronouns resolution problems that are trivial for humans but require commonsense reasoning for machines to solve correctly. The dataset consists of pairs of sentences differing in one or two words with ambiguous pronouns resolved differently in the two sentences, designed to test a system's commonsense reasoning abilities.

\textbf{CommonSenseQA (CSQA)} is a multiple-choice question answering dataset containing 12,102 questions with one correct answer and four distractor answers, requiring different types of commonsense knowledge to predict the correct answers. The dataset was constructed using ConceptNet relations and crowd-sourced questions to test commonsense reasoning.

\textbf{BoolQ} provides 15,942 yes/no questions that occur naturally in unconstrained environments, testing models' binary decision-making abilities.

\textbf{MMLU (Massive Multitask Language Understanding) } evaluates models across 57 diverse subjects covering STEM, humanities, and social sciences. The benchmark tests knowledge and problem-solving ability with content ranging from elementary to professional levels. This benchmark has become a standard evaluation metric in the field, with scores prominently reported for virtually all language models, and uses multiple-choice questions that allow for simple accuracy calculations.

\textbf{CMMLU (Chinese Massive Multitask Language Understanding) (CMLU)} Developed to address the gap in evaluating knowledge and reasoning capabilities in Chinese, CMMLU is a comprehensive benchmark covering 67 subjects from elementary to advanced professional levels across natural sciences, social sciences, engineering, and humanities. The benchmark includes topics with Chinese-specific answers that may not be universally applicable in other regions or languages, making it a fully Chinese-oriented evaluation tool. 

\textbf{RACE (Reading Comprehension from Examinations)} is collected from English examinations in China designed for middle and high school students, providing a culturally diverse reading assessment.

\textbf{XSum} evaluates abstract single document summarization systems, focusing on the ability to create concise one-sentence summaries capturing the essence of articles.

\textbf{C3 (Chinese Multiple-Choice Machine Reading Comprehension)} consists of multiple-choice questions from Chinese proficiency exams and ethnic Chinese exams.

\section{Task Arithmetic Merging}\label{sec:c}
Task Arithmetic \cite{ilharco2022editing} enhances model capabilities through vector operations by leveraging weighted combinations of task-specific knowledge. Given a base model with weights $\theta_{\text{pre}}$ and task-specific fine-tuned weights $\{\theta_{t}^{\text{ft}}\}_{t=1}^n$, task vectors are defined as $\tau_t = \theta_{t}^{\text{ft}} - \theta_{\text{pre}}$. The merged weights are then computed through $\theta_{\text{Merge}} = \theta_{\text{pre}} + \lambda \sum_{t=1}^n \tau_t$, where $\lambda$ controls the magnitude of task-specific adaptations.






\section{Full Baseline results} \label{sec:d}
To validate the efficiency of our proposed method, we conducted comparative experiments against established baseline techniques. For fair comparison with other baseline methods, we selected the same pruning ratios matching those used in LaCo \cite{yang2024laco} and ShortGPT \cite{men2024shortgpt} while being lower than those of other approaches. In order to make a fairer comparison, we reproduced all the results and evaluated them on OpenCompass \cite{2023opencompass} as in LaCo.All experiments run on NVIDIA Tesla A100 GPUs. For each baseline method, we explored three scenarios: (1) applying each baseline pruning method individually to all candidate models, (2) first pruning each candidate model using existing methods and then merging them, and (3) first merging the candidate models and then applying pruning techniques.

We use the official implement of LLM-pruner and LaCo, It's worth noting that when reproducing the LaCo method, we referenced the hyperparameter settings from the original paper. Due to differences in hardware, we couldn't fully reproduce the paper's results: we couldn't obtain models with pruning ratios consistent with the paper using the provided hyperparameters. We maintained consistency in all other parameters while gradually adjusting the threshold from 0.75 until achieving the desired pruning ratio. The specific parameters are detailed in the \Cref{tab:lacofig}.

For the reproduction of ShortGPT, we implemented the algorithm based on the original paper and similarly sampled 10,000 instances from the PG19 \cite{rae2019compressive} dataset as calibration data, following the methodology described in the paper. The resulting removed layers are shown in the Table. The removed layers for the base model align with those reported in the ShortGPT paper, albeit in a different sequence. We attribute this variation to slight differences in calculated layer importance scores. The specific configuration of removed layers for each model is detailed in the \Cref {tab:shortfig}.

For the merging process, we employed task arithmetic with weighting parameters in the range of [0.5, 1.0]. The full results of the baseline methods on the 7B model and the 13B model are presented in \Cref{tab:7bfullres} and \Cref {tab:13bfullres}, respectively.

\begin{table}[H]
  \centering
   \caption{Hyperparameter settings for LaCo results. $\mathcal{C}$: Number of layers combined in each merge; $\mathcal{L}$,$\mathcal{H}$: Layer range [$\mathcal{L}$, $\mathcal{H}$]; $\mathcal{I}$: Minimum interval between two adjacent merged layers; $\mathcal{T}$: Threshold for representation similarity.}
   \label{tab:lacofig}
  \renewcommand{\arraystretch}{1.2} 
  \begin{tabular}{c|l|ccccc}
    \toprule
    \textbf{Size} & \textbf{Model} & $\boldsymbol{\mathcal{C}}$ & $\boldsymbol{\mathcal{L}}$ & $\boldsymbol{\mathcal{H}}$ & $\boldsymbol{\mathcal{I}}$ & $\boldsymbol{\mathcal{T}}$ \\
    \midrule
    \multirow{8}{*}{Llama2-13B} 
    & Llama-2-13B & 6 & 1 & 40 & 2 & 0.7 \\
    & WizardLM-13B & 6 & 1 & 40 & 2 & 0.65 \\
    & WizardMath-13B & 6 & 1 & 40 & 2 & 0.7 \\
    & llama-2-13b-code-alpaca & 6 & 1 & 40 & 2 & 0.7 \\
    & Merge-then-prune & 6 & 1 & 40 & 2 & 0.65 \\
    & Prune-then-merge & 6 & 1 & 40 & 2 & 0.65 \\
    \midrule
    \multirow{8}{*}{Llama2-7B}
    & Llama-2-7B & 6 & 1 & 40 & 2 & 0.7 \\
    & Llama-2-7B-Chat & 6 & 1 & 40 & 2 & 0.65 \\
    & MAmmoTH-7B & 6 & 1 & 40 & 2 & 0.7 \\
    & Llama-2-Coder-7B & 6 & 1 & 40 & 2 & 0.7 \\
    & Merge-then-prune & 6 & 1 & 40 & 2 & 0.65 \\
    & Prune-then-merge & 6 & 1 & 40 & 2 & 0.65 \\
    \bottomrule
  \end{tabular}
\end{table}

\begin{table}[H]
    \centering
    \caption{Setup of Removed Layers for Candidate Models in ShortGPT.}
     \label{tab:shortfig}
    \begin{tabular}{l@{\hspace{2em}}l}
        \toprule
        \textbf{Model} & \textbf{Removed Layers} \\
        \midrule
        Llama-2-7B & 25, 27, 24, 28, 26, 29, 23, 22, 21 \\
        \addlinespace[0.5ex]
        Llama-2-7B-Chat & 27, 25, 24, 28, 29, 26, 23, 22, 21 \\
        \addlinespace[0.5ex]
        MAmmoTH-7B & 27, 25, 24, 28, 29, 23, 26, 22, 21 \\
        \addlinespace[0.5ex]
        Llama-2-Coder-7B & 27, 25, 24, 28, 29, 26, 23, 21, 22 \\
        \addlinespace[0.5ex]
        Llama-2-13B & 33, 32, 31, 30, 34, 35, 29, 28, 27, 26 \\ 
        \addlinespace[0.5ex]
        WizardLM-13B & 33, 32, 31, 30, 34, 35, 29, 28, 27, 36 \\
        \addlinespace[0.5ex]
        WizardMath-13B & 33, 31, 32, 30, 34, 35, 29, 28, 27, 36 \\
        \addlinespace[0.5ex]
         llama-2-13b-code-alpaca & 33, 31, 32, 30, 34, 35, 29, 28, 27, 26 \\
        \bottomrule
    \end{tabular}
\end{table}

\begin{table}[H]
\centering
\scriptsize
\caption{The main results of baseline methods on the 7B model across multiple natural language benchmarks using candidate models: Llama-2-7B-Chat (LM), MAmmoTH-7B (MAth), Llama-2-Coder-7B (Code), and Llama-2-7B (base).  "PTM" (Pruning-then-Merging) refers to first pruning each candidate model using current pruner and then merging them. "MTP" (Merging-then-Pruning) refers to first merging the candidate models and then applying pruning. For LLMPruner and SliceGPT, alignment challenges exist after pruning. LLMPruner removes different model blocks, while SliceGPT calculates orthogonal transformation matrices that are highly dependent on each model's specific weight distributions and activation patterns, resulting in incompatible transformation spaces. Therefore, we only implemented "merge then prune".}
\setlength{\tabcolsep}{0.8pt} 
\renewcommand{\arraystretch}{1.3} 

\definecolor{lightgray}{rgb}{0.9,0.9,0.9}
\definecolor{lightblue}{rgb}{0.8,0.85,1.0}

\begin{tabular}{ccc|>{\columncolor{white}}c>{\columncolor{white}}c>{\columncolor{white}}c|>{\columncolor{white}}c>{\columncolor{white}}c>{\columncolor{white}}c|>{\columncolor{white}}c>{\columncolor{white}}c>{\columncolor{white}}c>{\columncolor{white}}c|>{\columncolor{white}}c>{\columncolor{white}}c>{\columncolor{white}}c>{\columncolor{white}}c|>{}c}
\hline\hline
LLM & Pruner & Type & \multicolumn{3}{c|}{Reasoning} & \multicolumn{3}{c|}{Language} & \multicolumn{4}{c|}{Knowledge} & \multicolumn{4}{c|}{Understanding} & \multirow{2}{*}{Avg} \\
 & (ratio/layer) &  & CMNLI & HeSw & PIQA & CHID & WSC$_P$ & WSC$_G$ & CSQA & BoolQ & MMLU & CMLU & Race$_H$ & Race$_M$ & XSum & C3 & \\
\hline
\multirow{24}{*}{\makecell{Llama\\-7B}} & \multirow{4}{*}{Dense} & \cellcolor{lightgray}Base & \cellcolor{lightgray}32.98  & \cellcolor{lightgray}71.34 & \cellcolor{lightgray}78.18 & \cellcolor{lightgray}41.56 & \cellcolor{lightgray}37.50 & \cellcolor{lightgray}38.46 & \cellcolor{lightgray}55.04 & \cellcolor{lightgray}70.70 & \cellcolor{lightgray}46.67 & \cellcolor{lightgray}31.88 & \cellcolor{lightgray}35.53 & \cellcolor{lightgray}33.36 & \cellcolor{lightgray}19.55 & \cellcolor{lightgray}43.84 & \cellcolor{lightgray}45.47\\
& & \cellcolor{lightgray}Math & \cellcolor{lightgray}32.99 & \cellcolor{lightgray}68.60 & \cellcolor{lightgray}75.79 & \cellcolor{lightgray}39.71 & \cellcolor{lightgray}39.42 & \cellcolor{lightgray}36.54 & \cellcolor{lightgray}50.78 & \cellcolor{lightgray}69.36 & \cellcolor{lightgray}43.04 & \cellcolor{lightgray}32.16 & \cellcolor{lightgray}30.36 & \cellcolor{lightgray}36.42 & \cellcolor{lightgray}20.88 & \cellcolor{lightgray}43.45 & \cellcolor{lightgray}44.25\\
& & \cellcolor{lightgray}LM & \cellcolor{lightgray}31.30 & \cellcolor{lightgray}71.28 & \cellcolor{lightgray}75.95 & \cellcolor{lightgray}36.11 & \cellcolor{lightgray}63.46 & \cellcolor{lightgray}59.62 & \cellcolor{lightgray}64.29 & \cellcolor{lightgray}74.77 & \cellcolor{lightgray}48.30 & \cellcolor{lightgray}33.93 & \cellcolor{lightgray}52.52 & \cellcolor{lightgray}55.22 & \cellcolor{lightgray}22.45 & \cellcolor{lightgray}47.56 & \cellcolor{lightgray}52.63\\
& & \cellcolor{lightgray}Code & \cellcolor{lightgray}32.99 & \cellcolor{lightgray}70.27 & \cellcolor{lightgray}78.62 & \cellcolor{lightgray}41.61 & \cellcolor{lightgray}36.54 & \cellcolor{lightgray}41.35 & \cellcolor{lightgray}57.41 & \cellcolor{lightgray}71.04 & \cellcolor{lightgray}46.22 & \cellcolor{lightgray}32.20 & \cellcolor{lightgray}41.25 & \cellcolor{lightgray}39.69 & \cellcolor{lightgray}18.79 & \cellcolor{lightgray}46.25 & \cellcolor{lightgray}46.73\\
\cline{2-18}

& \multirow{5}{*}{\makecell{LLMPruner\\(25.32\%)}} & Base & 33.00 & 58.72 & 72.25 & 29.52 & 41.35 & 0.00 & 19.74 & 57.25 & 23.69 & 25.49 & 22.07 & 21.10 & 14.67 & 28.11 & 31.93\\
& & LM & 34.94 & 59.25 & 72.85 & 22.28 & 43.27 & 9.62 & 19.41 & 57.61 & 23.77 & 24.51 & 21.78 & 22.42 & 16.32 & 28.66 & 32.62\\
& & MATH & 32.99 & 55.74 & 70.84 & 25.82 & 37.50 & 21.15 & 18.84 & 54.31 & 24.77 & 25.20 & 22.87 & 23.89 & 10.91 & 28.00 & 32.35\\
& &\cellcolor{lightgray} Code & \cellcolor{lightgray}32.99 & \cellcolor{lightgray}59.57 & \cellcolor{lightgray}73.34 & \cellcolor{lightgray}30.32 & \cellcolor{lightgray}46.15 & \cellcolor{lightgray}0.00 & \cellcolor{lightgray}20.15 & \cellcolor{lightgray}57.28 & \cellcolor{lightgray}23.21 & \cellcolor{lightgray}25.16 & \cellcolor{lightgray}21.56 & \cellcolor{lightgray}21.52 & \cellcolor{lightgray}15.19 & \cellcolor{lightgray}31.07 & \cellcolor{lightgray}32.68\\
& & \cellcolor{lightblue}MTP & \cellcolor{lightblue}34.71 & \cellcolor{lightblue}60.57 & \cellcolor{lightblue}73.50 & \cellcolor{lightblue}26.62 & \cellcolor{lightblue}40.38 & \cellcolor{lightblue}5.77 & \cellcolor{lightblue}19.90 & \cellcolor{lightblue}52.14 & \cellcolor{lightblue}24.01 & \cellcolor{lightblue}25.30 & \cellcolor{lightblue}23.07 & \cellcolor{lightblue}22.98 & \cellcolor{lightblue}15.51 & \cellcolor{lightblue}32.49 & \cellcolor{lightblue}32.64\\
\cline{2-18}

& \multirow{5}{*}{\makecell{SliceGPT\\(26.33\%)}} & Base & 31.08 & 42.90 & 61.43 & 19.53 & 36.54 & 0.00 & 20.88 & 37.95 & 24.78 & 24.78 & 21.24 & 21.73 & 6.58 & 37.42 & 27.63\\
& & LM & 31.70 & 43.50 & 61.37 & 18.28 & 40.38 & 0.96 & 21.21 & 38.96 & 25.56 & 25.28 & 21.93 & 22.42 & 13.13 & 38.36 & 28.79\\
& & \cellcolor{lightgray}MATH & \cellcolor{lightgray}31.89 & \cellcolor{lightgray}41.55 & \cellcolor{lightgray}58.81 & \cellcolor{lightgray}18.43 & \cellcolor{lightgray}39.42 & \cellcolor{lightgray}4.81 & \cellcolor{lightgray}19.49 & \cellcolor{lightgray}40.09 & \cellcolor{lightgray}25.38 & \cellcolor{lightgray}25.02 & \cellcolor{lightgray}25.59 & \cellcolor{lightgray}26.88 & \cellcolor{lightgray}8.78 & \cellcolor{lightgray}39.56 & \cellcolor{lightgray}28.98\\
& & Code & 31.81 & 44.02 & 63.17 & 18.48 & 36.54 & 13.46 & 19.74 & 37.92 & 24.71 & 25.22 & 21.41 & 21.66 & 2.59 & 38.19 & 28.49\\
& & \cellcolor{lightblue}MTP & \cellcolor{lightblue}32.85 & \cellcolor{lightblue}37.61 & \cellcolor{lightblue}57.56 & \cellcolor{lightblue}17.33 & \cellcolor{lightblue}53.85 & \cellcolor{lightblue}2.88 & \cellcolor{lightblue}19.41 & \cellcolor{lightblue}42.66 & \cellcolor{lightblue}25.22 & \cellcolor{lightblue}24.68 & \cellcolor{lightblue}25.21 & \cellcolor{lightblue}24.72 & \cellcolor{lightblue}12.78 & \cellcolor{lightblue}40.22 & \cellcolor{lightblue}29.78\\
\cline{2-18}

& \multirow{6}{*}{LACO} & Base & 32.85 & 53.33 & 68.23 & 31.62 & 36.54 & 4.81 & 20.39 & 62.02 & 26.60 & 25.27 & 24.70 & 23.61 & 9.38 & 42.47 & 32.99\\
& & \cellcolor{lightgray}LM & \cellcolor{lightgray}32.97 & \cellcolor{lightgray}55.24 & \cellcolor{lightgray}69.53 & \cellcolor{lightgray}31.47 & \cellcolor{lightgray}36.54 & \cellcolor{lightgray}34.62 & \cellcolor{lightgray}22.11 & \cellcolor{lightgray}67.22 & \cellcolor{lightgray}29.08 & \cellcolor{lightgray}26.16 & \cellcolor{lightgray}28.53 & \cellcolor{lightgray}28.27 & \cellcolor{lightgray}14.68 & \cellcolor{lightgray}43.51 & \cellcolor{lightgray}37.14\\
& & Math & 32.97 & 55.24 & 69.53 & 31.47 & 50.00 & 34.62 & 22.11 & 67.22 & 29.44 & 26.16 & 22.53 & 23.68 & 14.68 & 39.34 & 37.07\\
& & Code & 32.28 & 53.68 & 69.15 & 32.22 & 36.54 & 1.92 & 20.56 & 61.99  & 26.31 & 25.43 & 27.10 & 22.70 & 11.14 & 43.07 & 33.15\\
& & MTP & 32.43 & 57.80 & 71.82 & 28.97 & 41.35 & 16.35 & 27.52 & 71.28  & 30.49 & 26.88 & 25.76 & 27.09 & 8.27 & 44.33 & 36.45\\
& &  \cellcolor{lightblue}PTM &  \cellcolor{lightblue}31.89 &  \cellcolor{lightblue}56.26 &  \cellcolor{lightblue}71.22 &  \cellcolor{lightblue}27.32 &  \cellcolor{lightblue}39.42 &  \cellcolor{lightblue}22.12 &  \cellcolor{lightblue}23.42 &  \cellcolor{lightblue}72.66  &  \cellcolor{lightblue}29.30 &  \cellcolor{lightblue} \cellcolor{lightblue}26.00 &  \cellcolor{lightblue}25.19 &  \cellcolor{lightblue}26.81 &  \cellcolor{lightblue}16.11 &  \cellcolor{lightblue}43.62 &  \cellcolor{lightblue}36.52\\
\cline{2-18}

& \multirow{6}{*}{\makecell{ShortGPT\\(27.1\%)}} & \cellcolor{lightgray}Base & \cellcolor{lightgray}33.09 & \cellcolor{lightgray}57.42 & \cellcolor{lightgray}66.54 & \cellcolor{lightgray}21.53 & \cellcolor{lightgray}56.73 & \cellcolor{lightgray}48.08 & \cellcolor{lightgray}52.5 & \cellcolor{lightgray}67.34 & \cellcolor{lightgray}43.68 & \cellcolor{lightgray}28.31 & \cellcolor{lightgray}32.53 & \cellcolor{lightgray}31.69 & \cellcolor{lightgray}12.40 & \cellcolor{lightgray}39.45 & \cellcolor{lightgray}42.24\\
& & LM & 33.85 & 53.93 & 63.82 & 14.59 & 39.42 & 22.12 & 58.48 & 67.95 & 35.85 & 26.60 & 48.03 & 51.18 & 6.93 & 37.21 & 40.00\\
& & MATH & 33.97 & 56.69 & 63.38 & 17.78 & 54.81 & 44.23 & 37.26 & 69.82 & 30.68 & 25.26 & 28.24 & 30.29 & 8.26 & 31.67 & 38.02\\
& & Code & 32.74 & 56.69 & 65.07 & 17.78 & 58.65 & 35.58 & 53.24 & 67.52 & 44.82 & 28.92 & 35.62 & 37.53 & 14.32 & 40.66 & 42.08\\
& & \cellcolor{lightblue}MTP & \cellcolor{lightblue}34.10 & \cellcolor{lightblue}54.18 & \cellcolor{lightblue}64.42 & \cellcolor{lightblue}16.83 & \cellcolor{lightblue}61.54 & \cellcolor{lightblue}36.54 & \cellcolor{lightblue}55.61 & \cellcolor{lightblue}73.21 & \cellcolor{lightblue}36.84 & \cellcolor{lightblue}25.61 & \cellcolor{lightblue}42.94 & \cellcolor{lightblue}45.89 & \cellcolor{lightblue}10.12 & \cellcolor{lightblue}35.73 & \cellcolor{lightblue}42.40\\
& & PTM & 34.10 & 54.18 & 64.42 & 16.83 & 61.54 & 36.54 & 55.61 & 73.21 & 36.84 & 25.61 & 42.94 & 45.89 & 10.12 & 35.73 & 42.40\\
\hline\hline
\end{tabular}
\label{tab:7bfullres}
\end{table}

\begin{table}[H]
\centering
\scriptsize
\caption{The main results of baseline methods on the 13B model across multiple natural language benchmarks using candidate models: WizardLM-13B (LM), WizardMath-13B (Math), llama-2-13b-code-alpaca (Code), and Llama-2-13B (Base). "PTM" (Pruning-then-Merging) refers to first pruning each candidate model using the current pruner and then merging them. "MTP" (Merging-then-Pruning) refers to first merging the candidate models and then applying pruning. For LLMPruner and SliceGPT, alignment challenges exist after pruning. LLMPruner removes different model blocks, while SliceGPT calculates orthogonal transformation matrices that are highly dependent on each model's specific weight distributions and activation patterns, resulting in incompatible transformation spaces. Therefore, we only implemented "merge then prune"}
\setlength{\tabcolsep}{0.8pt} 
\renewcommand{\arraystretch}{1.3} 
\definecolor{lightgray}{rgb}{0.9,0.9,0.9}
\definecolor{lightblue}{rgb}{0.8,0.85,1.0}

\begin{tabular}{ccc|>{\columncolor{white}}c>{\columncolor{white}}c>{\columncolor{white}}c|>{\columncolor{white}}c>{\columncolor{white}}c>{\columncolor{white}}c|>{\columncolor{white}}c>{\columncolor{white}}c>{\columncolor{white}}c>{\columncolor{white}}c|>{\columncolor{white}}c>{\columncolor{white}}c>{\columncolor{white}}c>{\columncolor{white}}c|>{}c}
\hline\hline
LLM & Pruner & Type & \multicolumn{3}{c|}{Reasoning} & \multicolumn{3}{c|}{Language} & \multicolumn{4}{c|}{Knowledge} & \multicolumn{4}{c|}{Understanding} & \multirow{2}{*}{Avg} \\
 & ratio/layer &  & CMNLI & HeSw & PIQA & CHID & WSC$_P$ & WSC$_G$ & CSQA & BoolQ & MMLU & CMLU & Race$_H$ & Race$_M$ & XSum & C3 & \\
\hline
\multirow{24}{*}{\makecell{Llama\\-13B}} & \multirow{4}{*}{Dense} & \cellcolor{lightgray}Base & \cellcolor{lightgray}32.99  & \cellcolor{lightgray}74.77 & \cellcolor{lightgray}79.71 & \cellcolor{lightgray}47.35 & \cellcolor{lightgray}50.96 & \cellcolor{lightgray}63.46 & \cellcolor{lightgray}67.24 & \cellcolor{lightgray}71.38 & \cellcolor{lightgray}55.84 & \cellcolor{lightgray}38.74 & \cellcolor{lightgray}57.98 & \cellcolor{lightgray}60.17 & \cellcolor{lightgray}23.47 & \cellcolor{lightgray}47.51 & \cellcolor{lightgray}55.11\\
& & \cellcolor{lightgray}LM & \cellcolor{lightgray}35.36 & \cellcolor{lightgray}70.41 & \cellcolor{lightgray}78.73 & \cellcolor{lightgray}36.21 & \cellcolor{lightgray}57.69 & \cellcolor{lightgray}60.58 & \cellcolor{lightgray}65.03 & \cellcolor{lightgray}73.70 & \cellcolor{lightgray}53.48 & \cellcolor{lightgray}30.85 & \cellcolor{lightgray}66.12 & \cellcolor{lightgray}71.66 & \cellcolor{lightgray}22.44 & \cellcolor{lightgray}52.00 & \cellcolor{lightgray}55.30\\
& & \cellcolor{lightgray}MATH & \cellcolor{lightgray}32.99 & \cellcolor{lightgray}68.78 & \cellcolor{lightgray}77.26 & \cellcolor{lightgray}44.36 & \cellcolor{lightgray}36.54 & \cellcolor{lightgray}19.23 & \cellcolor{lightgray}60.36 & \cellcolor{lightgray}78.44 & \cellcolor{lightgray}54.21 & \cellcolor{lightgray}38.12 & \cellcolor{lightgray}47.74 & \cellcolor{lightgray}48.82 & \cellcolor{lightgray}19.51 & \cellcolor{lightgray}44.66 & \cellcolor{lightgray}47.93\\
& & \cellcolor{lightgray}Code & \cellcolor{lightgray}32.99 & \cellcolor{lightgray}74.82 & \cellcolor{lightgray}80.14 & \cellcolor{lightgray}47.30 & \cellcolor{lightgray}51.92 & \cellcolor{lightgray}63.46 & \cellcolor{lightgray}68.88 & \cellcolor{lightgray}72.72 & \cellcolor{lightgray}55.92 & \cellcolor{lightgray}39.26 & \cellcolor{lightgray}58.03 & \cellcolor{lightgray}63.72 & \cellcolor{lightgray}24.45 & \cellcolor{lightgray}48.38 & \cellcolor{lightgray}55.86\\
\cline{2-18}

& \multirow{5}{*}{\makecell{LLMPruner\\(21.2\%)}} & Base & 33.27 & 63.57 & 75.41 & 34.17 & 37.50 & 0.00 & 19.57 & 45.35 & 23.08 & 25.36 & 21.61 & 21.80 & 14.41 & 29.64 & 31.77\\
& & \cellcolor{lightgray}LM & \cellcolor{lightgray}33.49 & \cellcolor{lightgray}60.28 & \cellcolor{lightgray}75.57 & \cellcolor{lightgray}23.68 & \cellcolor{lightgray}39.42 & \cellcolor{lightgray}0.00 & \cellcolor{lightgray}19.00 & \cellcolor{lightgray}63.24 & \cellcolor{lightgray}23.27 & \cellcolor{lightgray}25.23 & \cellcolor{lightgray}22.36 & \cellcolor{lightgray}21.45 & \cellcolor{lightgray}17.13 & \cellcolor{lightgray}32.00 & \cellcolor{lightgray}32.58\\
& & MATH & 32.99 & 55.49 & 72.91 & 30.02 & 41.35 & 0.00 & 19.08 & 53.18 & 23.06 & 25.53 & 21.36 & 21.31 & 12.25 & 29.10 & 31.26\\
& & Code & 33.18 & 64.21 & 75.52 & 34.17 & 43.27 & 0.00 & 19.90 & 47.80 & 23.19 & 25.52 & 21.61 & 22.08 & 16.08 & 29.59 & 32.58\\
& & \cellcolor{lightblue}MTP & \cellcolor{lightblue}33.86 & \cellcolor{lightblue}64.11 & \cellcolor{lightblue}73.50 & \cellcolor{lightblue}22.18 & \cellcolor{lightblue}60.58 & \cellcolor{lightblue}0.00 & \cellcolor{lightblue}21.46 & \cellcolor{lightblue}61.96 & \cellcolor{lightblue}23.84 & \cellcolor{lightblue}25.62 & \cellcolor{lightblue}22.16 & \cellcolor{lightblue}21.59 & \cellcolor{lightblue}14.98 & \cellcolor{lightblue}32.11 & \cellcolor{lightblue}34.14\\
\cline{2-18}

& \multirow{5}{*}{\makecell{SliceGPT\\(23.6\%)}} & Base & 30.39 & 46.69 & 63.22 & 18.78 & 42.31 & 25.96 & 25.23 & 37.83 & 30.43 & 25.14 & 23.47 & 24.65 & 8.78 & 39.56 & 31.60\\
& & \cellcolor{lightgray}LM & \cellcolor{lightgray}33.19 & \cellcolor{lightgray}42.44 & \cellcolor{lightgray}59.90 & \cellcolor{lightgray}18.03 & \cellcolor{lightgray}54.81 & \cellcolor{lightgray}19.23 & \cellcolor{lightgray}32.51 & \cellcolor{lightgray}41.22 & \cellcolor{lightgray}33.09 & \cellcolor{lightgray}25.75 & \cellcolor{lightgray}29.45 & \cellcolor{lightgray}29.87 & \cellcolor{lightgray}9.99 & \cellcolor{lightgray}37.75 & \cellcolor{lightgray}33.37\\
& & MATH & 32.73 & 36.27 & 59.30 & 17.38 & 42.31 & 0.00 & 21.62 & 37.83 & 30.33 & 25.16 & 23.84 & 24.16 & 1.54 & 40.82 & 28.09\\
& & Code & 30.82 & 46.69 & 63.00 & 19.18 & 42.31 & 27.88 & 24.82 & 37.83 & 31.38 & 25.20 & 23.47 & 24.65 & 8.83 & 40.00 & 31.86\\
& & \cellcolor{lightblue}MTP & \cellcolor{lightblue}30.98 & \cellcolor{lightblue}46.83 & \cellcolor{lightblue}62.57 & \cellcolor{lightblue}19.33 & \cellcolor{lightblue}51.92 & \cellcolor{lightblue}49.04 & \cellcolor{lightblue}37.76 & \cellcolor{lightblue}38.38 & \cellcolor{lightblue}33.55 & \cellcolor{lightblue}25.22 & \cellcolor{lightblue}23.53 & \cellcolor{lightblue}23.05 & \cellcolor{lightblue}9.95 & \cellcolor{lightblue}39.67 & \cellcolor{lightblue}35.13\\
\cline{2-18}

& \multirow{6}{*}{\makecell{LaCo\\(24.6\%)}} & Base & 32.97 & 59.38 & 73.45 & 36.26 & 37.50 & 37.50 & 19.41 & 57.31 & 25.03 & 24.41 & 22.47 & 23.19 & 16.39 & 37.92 & 35.94\\
& & \cellcolor{lightgray}LM & \cellcolor{lightgray}32.33 & \cellcolor{lightgray}60.18 & \cellcolor{lightgray}70.57 & \cellcolor{lightgray}32.67 & \cellcolor{lightgray}34.62 & \cellcolor{lightgray}34.62 & \cellcolor{lightgray}52.58 & \cellcolor{lightgray}62.66 & \cellcolor{lightgray}36.26 & \cellcolor{lightgray}25.80 & \cellcolor{lightgray}60.38 & \cellcolor{lightgray}62.53 & \cellcolor{lightgray}8.79 & \cellcolor{lightgray}49.21 & \cellcolor{lightgray}44.51\\
& & Math & 33.97 & 56.51 & 72.25 & 33.52 & 44.23 & 44.23 & 21.38 & 64.19 & 25.35 & 24.55 & 21.98 & 21.94 & 12.77 & 37.48 & 36.74\\
& & Code & 32.99 & 59.53 & 75.03 & 38.41 & 51.92 & 0.00 & 19.49 & 53.18  & 24.48 & 24.72 & 22.87 & 22.28 & 17.70 & 37.53 & 34.30\\
& & \cellcolor{lightblue}MTP & \cellcolor{lightblue}33.49 & \cellcolor{lightblue}62.50 & \cellcolor{lightblue}74.37 & \cellcolor{lightblue}35.26 & \cellcolor{lightblue}63.46 & \cellcolor{lightblue}63.46 & \cellcolor{lightblue}18.84 & \cellcolor{lightblue}64.65  & \cellcolor{lightblue}41.83 & \cellcolor{lightblue}24.87 & \cellcolor{lightblue}26.10 & \cellcolor{lightblue}25.97 & \cellcolor{lightblue}15.93 & \cellcolor{lightblue}39.51 &\cellcolor{lightblue}42.16 \\
& & PTM & 31.85 & 29.80 & 51.31 & 12.74 & 36.54 & 36.54 & 19.57 & 62.08  & 24.37 & 25.19 & 22.10 & 22.77 & 0.40 & 35.12 & 29.31\\
\cline{2-18}

& \multirow{6}{*}{\makecell{ShortGPT\\(24.6\%)}} & Base & 32.99 & 67.07 & 73.45 & 36.46 & 42.31 & 45.19 & 66.99 & 58.56 & 54.74 & 38.39 & 56.89 & 54.06 & 18.58 & 46.19 & 49.42\\
& & \cellcolor{lightgray}LM & \cellcolor{lightgray}32.95 & \cellcolor{lightgray}62.64 & \cellcolor{lightgray}73.50 & \cellcolor{lightgray}28.22 & \cellcolor{lightgray}36.54 & \cellcolor{lightgray}50.96 & \cellcolor{lightgray}65.44 & \cellcolor{lightgray}67.71 & \cellcolor{lightgray}53.50 & \cellcolor{lightgray}30.73 & \cellcolor{lightgray}65.52 & \cellcolor{lightgray}71.38 & \cellcolor{lightgray}19.12 & \cellcolor{lightgray}48.60 & \cellcolor{lightgray}50.49\\
& & MATH & 32.99 & 59.63 & 70.40 & 31.12 & 40.38 & 1.92 & 59.71 & 70.00 & 52.70 & 36.94 & 43.51 & 44.29 & 7.73 & 43.84 & 42.51 \\
& & Code & 32.92 & 67.03 & 74.37 & 36.41 & 55.77 & 46.15 & 68.96 & 60.55 & 54.94 & 38.30 & 53.60 & 58.57 & 8.41 & 47.18 & 50.23\\
& & \cellcolor{lightblue}MTP & \cellcolor{lightblue}31.07 & \cellcolor{lightblue}63.24 & \cellcolor{lightblue}68.61 & \cellcolor{lightblue}27.17 & \cellcolor{lightblue}49.04 & \cellcolor{lightblue}43.27 & \cellcolor{lightblue}65.68 & \cellcolor{lightblue}78.01 & \cellcolor{lightblue}51.26 & \cellcolor{lightblue}36.88 & \cellcolor{lightblue}57.38 & \cellcolor{lightblue}62.67 & \cellcolor{lightblue}16.94 & \cellcolor{lightblue}44.05 & \cellcolor{lightblue}49.66\\
& & PTM & 31.08 & 63.32 & 68.66 & 27.12 & 49.04 & 43.27 & 65.68 & 77.98 & 51.23 & 36.82 & 57.40 & 62.47 & 17.01 & 43.95 & 49.65\\
\hline\hline
\end{tabular}
\label{tab:13bfullres}
\end{table}

\definecolor{lightgray}{rgb}{0.9,0.9,0.9}
\definecolor{lightblue}{rgb}{0.8,0.85,1.0}
\setlength{\tabcolsep}{1.3pt} 
\renewcommand{\arraystretch}{1.2} 
\begin{table}[H]
\centering
\caption{Performance comparison of various model pruning strategies across multiple benchmark categories. The settings include LR-only (Layer Removal only), LS+LR (combined Layer Selection and Layer Removal), FL-merge (Folding Layers Merging), Single-obj (Single-objective optimization), and PPL-obj (Perplexity-based objective). For multi-objective optimization approaches, three representative Pareto-optimal solutions (numbered 1-3) are presented.}
\scriptsize
\begin{tabular}{l ccc ccc cccc cccc c}
\toprule
\rowcolor{gray!16}
\textbf{setting} & \multicolumn{3}{c}{\textbf{Reasoning}} & \multicolumn{3}{c}{\textbf{Language}} & \multicolumn{4}{c}{\textbf{Knowledge}} & \multicolumn{4}{c}{\textbf{Understanding}} & \textbf{Avg} \\
\cmidrule(lr){2-4} \cmidrule(lr){5-7} \cmidrule(lr){8-11} \cmidrule(lr){12-15}
\rowcolor{gray!16}
 & \textbf{CNLI} & \textbf{HeSw} & \textbf{PIQA} & \textbf{CHID} & \textbf{WSC$_P$} & \textbf{WSC$_G$} & \textbf{CSQA} & \textbf{BoolQ} & \textbf{MMLU} & \textbf{CMLU} & \textbf{Race$_H$} & \textbf{Race$_M$} & \textbf{XSum} & \textbf{C3} & \\
\midrule
LR-only-LM-1 &33.93 & 57.51& 65.49&18.18& 62.46& 48.03&58.79 & 62.18& 45.76&30.95 &49.54&53.36&1.45&38.60&44.73 \\
LR-only-LM-2 & 33.58&52.10&64.25 &19.53&50.00 &62.50&63.64&41.80 &48.33&32.84& 51.03& 51.46&5.47 &39.56& 44.01\\
LR-only-LM-3 &34.96 &53.80&66.70 &18.58&49.04 &58.65 &60.61&68.87&47.85&33.54&42.51&43.04&8.05 &41.42 &44.83\\
LR-only-Math-1 &33.77 &54.49&68.23 &21.93 &62.50 &37.50& 27.85& 57.52& 37.08&28.73& 31.42&34.05& 7.51&37.92 &38.61\\
LR-only-Math-2 & 31.69&56.56 & 68.77&27.07&63.46 &30.77&36.69 &62.35 &39.17 &29.15 &33.39 & 38.65&4.41 &43.34 & 40.39\\
LR-only-Math-3 &32.94 &58.43&69.64&25.97 &54.81&25.96 &29.89&62.84&33.46 &26.92 &31.39 & 32.10&8.06&40.16 & 38.04 \\
LR-only-Code-1 & 30.13 &57.60& 70.35&27.07 & 63.46& 11.54 & 50.94&65.96 &42.64 &30.96& 36.39&36.77 &3.15&43.78&40.77 \\
LR-only-Code-2 &34.94 &57.37&68.55&28.67&42.31&41.35&54.46&63.00&42.49 &27.39 &34.88 & 35.31&4.08&43.78& 41.33\\
LR-only-Code-3 &34.93 &56.71 &69.42 &25.92&59.62 &31.65 &52.83&62.20 &43.03&28.80&38.51&39.07 &2.87 & 41.70& 41.95\\
LR-only-Base-1 & 32.67 &54.21&66.00&26.07&36.54&1.92 & 49.47&64.19&44.47 &28.84&38.99&38.86 &0.25 &41.59 &37.43 \\
LR-only-Base-2 & 32.22 &56.48& 67.46&26.32&61.54 & 50.00& 41.44& 66.91& 40.54&28.01&37.94 &39.35 &0.96& 41.92&42.22 \\
LR-only-Base-3 &31.13&52.90 &67.95&27.97&36.54 &0.00 & 54.63& 64.13&43.01 &30.03 &35.56& 37.05&6.79 &41.70& 37.81\\
\midrule
FL-merge-1 &32.99 &52.90&63.66&19.28 &46.15&62.50&60.52 &75.20&48.30&34.33 &50.77 &55.29& 6.39&39.40&46.26 \\
FL-merge-2 &32.99 &51.99 & 63.44&18.33 &46.15&63.46&61.26&74.77& 48.80&33.84&51.11 &56.34&5.75 &37.86&46.15 \\
FL-merge-3 &33.89 &51.15 &62.62 &18.63&50.00&61.54& 60.44&75.78 & 48.61&33.96 &50.74& 55.85&5.72 &38.03 & 46.15\\
\midrule
LS+LR-1 &34.75 & 53.65&66.32 &17.83 & 63.46& 22.12& 59.71&70.61 &47.32&33.77&36.62 &33.91& 8.54 &42.35&42.21 \\
LS+LR-2 &31.74&55.25& 68.39&26.77 &63.46& 10.58&58.72&66.27& 47.40&33.15 &40.02 &45.26&2.62 &44.16&42.41\\
LS+LR-3 &32.92&55.84&65.07 &17.98 &63.46&26.92& 58.97&51.22& 48.97&34.61&48.68&49.44& 8.33&42.41& 43.20\\
\midrule
Single-obj &32.15 &56.02 &67.46&19.08&39.42 &48.08&62.33& 74.43&47.40& 34.14&50.94&52.86&12.35&41.97&45.62\\
PPL-obj &33.39&23.89 &52.07& 14.84& 45.19&7.69 &19.33&39.51 & 24.25&24.69 &22.81&21.17&0.06&26.36&25.38\\
\bottomrule
\end{tabular}
 \label{tab:fullablation}
\label{tab:model_comparisonfull}
\end{table}

\definecolor{lightgray}{rgb}{0.9,0.9,0.9}
\definecolor{lightblue}{rgb}{0.8,0.85,1.0}
\setlength{\tabcolsep}{1.3pt} 
\renewcommand{\arraystretch}{1.2} 
\begin{table}[H]
\centering
\caption{Model Performance Comparison Across Pruning Ratios}
\scriptsize
\begin{tabular}{l l ccc ccc cccc cccc c}
\toprule
\rowcolor{gray!16}
\textbf{Model} & \textbf{Prune Ratio} & \multicolumn{3}{c}{\textbf{Reasoning}} & \multicolumn{3}{c}{\textbf{Language}} & \multicolumn{4}{c}{\textbf{Knowledge}} & \multicolumn{4}{c}{\textbf{Understanding}} & \textbf{Avg} \\
\cmidrule(lr){3-5} \cmidrule(lr){6-8} \cmidrule(lr){9-12} \cmidrule(lr){13-16}
\rowcolor{gray!16}
 & & \textbf{CNLI} & \textbf{HeSw} & \textbf{PIQA} & \textbf{CHID} & \textbf{WSC$_P$} & \textbf{WSC$_G$} & \textbf{CSQA} & \textbf{BoolQ} & \textbf{MMLU} & \textbf{CMLU} & \textbf{Race$_H$} & \textbf{Race$_M$} & \textbf{XSum} & \textbf{C3} & \\
\midrule
Base & 0 & 32.98  & 71.34 & 78.18 & 41.56 & 37.50 & 38.46 & 55.04 & 70.70 & 46.67 & 31.88 & 35.53 & 33.36 & 19.55 & 43.84 & 45.47\\
Base & 12.5 & 32.99 & 67.06 & 74.92 & 39.61 & 36.53 & 1.92 & 57.41 & 69.36 & 47.15 & 31.61 & 39.11 & 38.65 & 17.59 & 44.60 & 42.75 \\

Base & 25 & 32.98 & 63.80 & 69.21 & 35.37 & 36.54 & 0.00 & 50.78 & 64.74 & 40.80 & 30.31 & 35.19 & 35.62 & 16.11 & 43.51& 39.64 \\
Base & 37.5 & 32.58 & 45.04 & 61.53 & 20.68 & 36.54 & 2.88 & 42.18 & 64.43 & 39.87 & 29.42 & 31.90 & 29.74 & 2.77 & 41.37 & 34.35 \\

Base & 50 & 34.51 & 34.89 & 55.33 & 17.08 & 36.54 & 11.54 & 19.82 & 62.29 & 28.72 & 25.10 & 23.41 & 26.04 & 1.21 & 35.07 & 29.40 \\

Base & 62.5 & 35.14 & 29.71 & 52.83 & 14.94 & 39.42 & 1.92 & 21.46 & 50.06 & 24.55 & 25.16 & 26.76 & 25.42 & 0.09 & 27.62 & 26.80 \\
Base & 75 & 34.94 & 26.71 & 51.03 & 13.59 & 36.54 & 8.65 & 20.56 & 52.60 & 24.23 & 24.47 & 23.18 & 22.63 & 0.08 & 27.29 & 26.17 \\
\midrule
LM & 0 & 31.30 & 71.28 & 75.95 & 36.11 & 63.46 & 59.62 & 64.29 & 74.77 & 48.30 & 33.93 & 52.52 & 55.22 & 22.45 & 47.56 & 52.63\\
LM & 12.5 & 32.42 & 67.58 & 72.72 & 28.91 & 50.92 & 60.50 & 60.92 & 72.88 & 46.69 & 32.02 & 51.34 & 54.45 & 18.26 & 45.94 & 49.68 \\
LM & 25 & 30.10 & 60.63 & 66.82 & 20.53 & 48.96 & 42.31 & 65.88 & 70.82 & 42.09 & 32.40 & 48.23 & 50.43 & 15.75 & 43.62 & 45.11 \\
LM & 37.5 & 33.29 & 45.13 & 60.66 & 20.03 & 36.54 & 11.73 & 59.38 & 68.07 & 39.18 & 29.64 & 39.71 & 42.20 & 6.36 & 41.04 & 39.40 \\
LM & 50 & 34.93 & 34.67 & 56.20 & 16.18 & 36.54 & 8.65 & 22.28 & 62.14 & 32.01 & 26.44 & 25.39 & 25.49 & 2.34 & 35.01 & 29.88 \\
LM & 62.5 & 34.11 & 30.50 & 53.21 & 14.34 & 51.92 & 2.88 & 20.56 & 57.95 & 24.58 & 25.21 & 23.13 & 23.75 & 0.18 & 27.12 & 27.82 \\
LM & 75 & 34.87 & 27.03 & 52.19 & 14.54 & 39.42 & 0.00 & 20.23 & 53.87 & 24.45 & 24.83 & 21.41 & 22.14 & 0.02 & 26.69 & 25.82 \\
\midrule

Math& 0 & 32.99 & 68.60 & 75.79 & 39.71 & 39.42 & 36.54 & 50.78 & 69.36 & 43.04 & 32.16 & 30.36 & 36.42 & 20.88 & 43.45 & 44.25\\

Math & 12.5 & 32.97 & 64.72 & 73.06 & 37.50 & 23.08 & 23.07 & 51.43 & 71.16 & 42.91 & 31.90 & 32.99 & 36.07 & 19.30 & 43.83 & 41.71 \\
Math & 25 & 34.92 & 46.24 & 61.92 & 19.38 & 36.54 & 56.73 & 45.45 & 72.81 & 35.07 & 29.78 & 31.45 & 34.33 & 6.24 & 39.89 & 39.34 \\
Math & 37.5 & 32.99 & 55.42 & 62.81 & 23.82 & 38.38 & 4.81 & 37.87 & 68.68 & 36.46 & 27.19 & 28.02 & 33.79 & 13.88 & 39.37 & 36.04 \\
Math & 50 & 32.73 & 35.93 & 55.06 & 16.73 & 39.42 & 39.42 & 20.15 & 64.34 & 29.94 & 25.52 & 26.82 & 26.60 & 2.31 & 35.56 & 32.15 \\
Math & 62.5 & 34.93 & 31.06 & 54.08 & 13.79 & 58.65 & 4.81 & 20.56 & 46.24 & 26.70 & 25.05 & 26.56 & 26.53 & 0.57 & 28.33 & 28.42 \\
Math & 75 & 34.94 & 27.35 & 52.07 & 14.39 & 43.27 & 2.88 & 20.88 & 56.51 & 24.25 & 23.14 & 24.76 & 24.79 & 0.15 & 27.45 & 27.20 \\
\midrule
Code & 0 & 32.99 & 70.27 & 78.62 & 41.61 & 36.54 & 41.35 & 57.41 & 71.04 & 46.22 & 32.20 & 41.25 & 39.69 & 18.79 & 46.25 & 46.73\\
Code & 12.5 & 32.97 & 65.79 & 75.78 & 39.06 & 36.54 & 0.96 & 56.67 & 71.13 & 47.09 & 32.00 & 44.73 & 44.84 & 19.21 & 47.29 & 43.86 \\
Code & 25 & 32.99 & 63.06 & 72.02 & 35.67 & 36.54 & 0.00 & 50.59 & 68.87 & 40.50 & 28.87 & 36.64 & 38.59 & 17.59 & 45.64 & 40.51 \\
Code & 37.5 & 33.21 & 44.12 & 62.13 & 20.78 & 36.54 & 2.88 & 48.81 & 63.91 & 40.29 & 29.56 & 36.25 & 35.52 & 5.35 & 42.14 & 35.82 \\
Code & 50 & 34.93 & 34.15 & 54.95 & 16.73 & 36.54 & 17.31 & 22.03 & 62.54 & 28.46 & 25.16 & 24.13 & 24.44 & 2.03 & 36.62 & 30.00 \\
Code & 62.5 & 34.72 & 29.67 & 52.99 & 14.39 & 40.38 & 8.65 & 22.52 & 50.70 & 24.78 & 25.15 & 27.16 & 28.04 & 0.12 & 27.78 & 27.50 \\
Code & 75 & 34.94 & 26.79 & 50.82 & 13.99 & 38.46 & 5.77 & 24.08 & 48.38 & 24.08 & 24.52 & 22.73 & 22.49 & 0.13 & 27.29 & 26.03 \\
\midrule
Ours & 0 & 36.88 & 73.16 & 78.67 & 39.46 & 64.46 & 45.19 & 65.37 & 78.43 & 49.75 & 35.08 & 58.78 & 61.65 & 24.50 & 49.33 & 54.34 \\
Ours & 12.5 & 33.00 & 66.78 & 75.19 & 34.92 & 64.42 & 63.46 & 63.98 & 75.87 & 48.79 & 34.13 & 53.89 & 56.20 & 20.21 & 45.37 & 52.59 \\
Ours & 25 & 32.99 & 57.31 & 68.34 & 22.38 & 63.46 & 63.46 & 57.58 & 62.17 & 45.92 & 30.96 & 52.20 & 56.06 & 7.12 & 39.67 & 47.11 \\
Ours & 37.5 & 35.67 & 51.02 & 63.44 & 20.68 & 62.50 & 22.00 & 57.99 & 67.52 & 47.09 & 34.11 & 44.00 & 46.38 & 2.96 & 39.34 & 42.00 \\
Ours & 50 & 33.97 & 41.99 & 58.16 & 21.08 & 38.54 & 24.12 & 26.52 & 46.03 & 32.32 & 28.30 & 28.99 & 28.88 & 6.30 & 36.11 & 32.23 \\
Ours & 62.5 & 33.30 & 28.34 & 51.96 & 18.09 & 46.15 & 6.88 & 23.88 & 45.81 & 26.41 & 26.95 & 28.73 & 28.72 & 5.09 & 28.47 & 28.48 \\
Ours & 75 & 34.93 & 30.45 & 49.18 & 20.48 & 39.54 & 10.81 & 21.98 & 45.29 & 25.28 & 24.68 & 26.30 & 26.93 & 0.46 & 28.38 & 27.47 \\
\bottomrule
\end{tabular}
\label{tab:prune_ratio}
\end{table}

\begin{table}[H]
\centering
\scriptsize
\caption{The main results of the Llama3-8B model across multiple natural language benchmarks using candidate models: Meta-Llama-3-8B-Instruct (LM), MathCoder2-Llama-3-8B (Math), Code-Llama-3-8B (Code), and Meta-Llama-3-8B (Base). "PTM" (Pruning-then-Merging) refers to first pruning each candidate model using the current pruner and then merging them. "MTP" (Merging-then-Pruning) refers to first merging the candidate models and then applying pruning.}
\setlength{\tabcolsep}{0.8pt} 
\renewcommand{\arraystretch}{1.3} 
\definecolor{lightgray}{rgb}{0.9,0.9,0.9}
\definecolor{lightblue}{rgb}{0.8,0.85,1.0}

\begin{tabular}{ccc|>{\columncolor{white}}c>{\columncolor{white}}c>{\columncolor{white}}c|>{\columncolor{white}}c>{\columncolor{white}}c>{\columncolor{white}}c|>{\columncolor{white}}c>{\columncolor{white}}c>{\columncolor{white}}c>{\columncolor{white}}c|>{\columncolor{white}}c>{\columncolor{white}}c>{\columncolor{white}}c>{\columncolor{white}}c|>{}c}
\hline\hline
LLM & Pruner & Type & \multicolumn{3}{c|}{Reasoning} & \multicolumn{3}{c|}{Language} & \multicolumn{4}{c|}{Knowledge} & \multicolumn{4}{c|}{Understanding} & \multirow{2}{*}{Avg} \\
 & ratio/layer &  & CMNLI & HeSw & PIQA & CHID & WSC$_P$ & WSC$_G$ & CSQA & BoolQ & MMLU & CMLU & Race$_H$ & Race$_M$ & XSum & C3 & \\
\hline
\multirow{8}{*}{\makecell{Llama3\\-8B}} & \multirow{4}{*}{Dense} & \cellcolor{lightgray}Base & \cellcolor{lightgray}32.98  & \cellcolor{lightgray}74.67 & \cellcolor{lightgray}80.96 & \cellcolor{lightgray}73.78 & \cellcolor{lightgray}56.73 & \cellcolor{lightgray}36.54 & \cellcolor{lightgray}73.79 & \cellcolor{lightgray}69.97 & \cellcolor{lightgray}64.74 & \cellcolor{lightgray}50.79 & \cellcolor{lightgray}63.21 & \cellcolor{lightgray}70.54 & \cellcolor{lightgray}3.28 & \cellcolor{lightgray}55.18 & \cellcolor{lightgray}57.65\\
& & \cellcolor{lightgray}LM & \cellcolor{lightgray}33.00 & \cellcolor{lightgray}71.08 & \cellcolor{lightgray}80.69 & \cellcolor{lightgray}65.53 & \cellcolor{lightgray}55.77 & \cellcolor{lightgray}69.23 & \cellcolor{lightgray}76.66 & \cellcolor{lightgray}78.87 & \cellcolor{lightgray}65.97 & \cellcolor{lightgray}53.64 & \cellcolor{lightgray}76.44 & \cellcolor{lightgray}81.75 & \cellcolor{lightgray}17.97 & \cellcolor{lightgray}63.95 & \cellcolor{lightgray}63.61\\
& & \cellcolor{lightgray}Math & \cellcolor{lightgray}32.99 & \cellcolor{lightgray}71.66 & \cellcolor{lightgray}77.97 & \cellcolor{lightgray}57.09 & \cellcolor{lightgray}37.50 & \cellcolor{lightgray}58.65 & \cellcolor{lightgray}68.22 & \cellcolor{lightgray}69.08 & \cellcolor{lightgray}62.08 & \cellcolor{lightgray}45.85 & \cellcolor{lightgray}64.75 & \cellcolor{lightgray}69.08 & \cellcolor{lightgray}8.68 & \cellcolor{lightgray}53.86 & \cellcolor{lightgray}55.53\\
& & \cellcolor{lightgray}Code & \cellcolor{lightgray}32.98 & \cellcolor{lightgray}65.56 & \cellcolor{lightgray}74.70 & \cellcolor{lightgray}78.42 & \cellcolor{lightgray}61.54 & \cellcolor{lightgray}61.54 & \cellcolor{lightgray}63.47 & \cellcolor{lightgray}78.35 & \cellcolor{lightgray}48.03 & \cellcolor{lightgray}34.55 & \cellcolor{lightgray}52.40 & \cellcolor{lightgray}58.43 & \cellcolor{lightgray}19.36 & \cellcolor{lightgray}46.41 & \cellcolor{lightgray}55.41\\
\cline{2-18}

& \multirow{6}{*}{\makecell{ShortGPT\\(24.6\%)}} & Base & 36.00 & 31.36 & 62.84 & 25.77 & 36.54 & 63.46 & 53.97 & 50.61 & 36.05 & 33.83 & 30.73 & 32.38 & 1.17 & 38.96 & 38.12\\
& & \cellcolor{lightgray}LM & \cellcolor{lightgray}32.83 & \cellcolor{lightgray}45.06 & \cellcolor{lightgray}65.78 & \cellcolor{lightgray}23.38 & \cellcolor{lightgray}41.35 & \cellcolor{lightgray}53.85 & \cellcolor{lightgray}39.56 & \cellcolor{lightgray}63.73 & \cellcolor{lightgray}32.37 & \cellcolor{lightgray}28.69 & \cellcolor{lightgray}40.14 & \cellcolor{lightgray}45.19 & \cellcolor{lightgray}3.68 & \cellcolor{lightgray}43.51 & \cellcolor{lightgray}39.94\\
& & Math & 32.98 & 42.89 & 63.00 & 17.18 & 36.54 & 36.54 & 45.37 & 46.30 & 33.95 & 29.71 & 28.87 & 30.22 & 1.45 & 40.49 & 34.68 \\
& & Code & 32.26 & 45.99 & 64.96 & 17.03 & 36.54 & 36.54 & 36.20 & 63.98 & 28.78 & 26.25 & 27.27 & 29.46 & 3.57 & 39.01 & 34.85\\
& & MTP & 32.98 & 48.51 & 64.85 & 18.33 & 36.54 & 35.58 & 42.83 & 67.06 & 33.05 & 28.73 & 30.07 & 32.66 & 3.64 & 44.33 & 37.08\\
& & \cellcolor{lightblue}PTM & \cellcolor{lightblue}32.95 & \cellcolor{lightblue}48.58 & \cellcolor{lightblue}64.96 & \cellcolor{lightblue}18.43 & \cellcolor{lightblue}36.54 & \cellcolor{lightblue}35.58 & \cellcolor{lightblue}42.83 & \cellcolor{lightblue}67.22 & \cellcolor{lightblue}33.05 & \cellcolor{lightblue}28.71 & \cellcolor{lightblue}30.16 & \cellcolor{lightblue}32.45 & \cellcolor{lightblue}3.66 & \cellcolor{lightblue}44.27 & \cellcolor{lightblue}37.10\\
\cline{2-18}

\hline\hline
\end{tabular}
\label{tab:8bfullres}
\end{table}

\begin{table}[H]
\centering
\caption{Architecture Parameters of pruned 13B models}
\resizebox{\textwidth}{!}{
\begin{tabular}{c|c|c|c|c|c|c|c|c|c}
\hline
\rowcolor{lightgray}
\multirow{1}{*}{\textbf{Layer}} & \multicolumn{3}{c|}{\textbf{Model-1}} & \multicolumn{3}{c|}{\textbf{Model-2}} & \multicolumn{3}{c}{\textbf{Model-3}} \\
\cline{2-10}
\rowcolor{lightgray}
 & \textbf{Type} & \begin{tabular}{c}\textbf{Merge}\\\textbf{Factor}\end{tabular} & \begin{tabular}{c}\textbf{Output}\\\textbf{Scale}\end{tabular} & \textbf{Type} & \begin{tabular}{c}\textbf{Merge}\\\textbf{Factor}\end{tabular} & \begin{tabular}{c}\textbf{Output}\\\textbf{Scale}\end{tabular} & \textbf{Type} & \begin{tabular}{c}\textbf{Merge}\\\textbf{Factor}\end{tabular} & \begin{tabular}{c}\textbf{Output}\\\textbf{Scale}\end{tabular} \\
\hline
0 & Base & - & 1.00 & LM & - & 1.00 & LM & - & 1.00 \\
\hline
1 & LM & - & 1.00 & LM+Math & 0.64 & 1.00 & Base & - & 1.00 \\
\hline
2 & LM & - & 1.00 & LM+Code & 0.60 & 1.05 & LM+Code & 0.60 & 1.05 \\
\hline
3 & LM & - & 1.00 & LM & - & 1.00 & LM+Code & 0.60 & 1.00 \\
\hline
4 & LM & - & 1.00 & LM & - & 1.00 & LM & - & 1.00 \\
\hline
5 & Code & - & 1.00 & LM+Math & 0.59 & 1.00 & LM+Math & 0.58 & 1.00 \\
\hline
6 & Base & - & 1.00 & LM & - & 1.00 & LM & - & 1.00 \\
\hline
7 & LM & - & 1.00 & LM+Math & 0.60 & 1.00 & LM+Math & 0.60 & 1.00 \\
\hline
8 & LM & - & 1.00 & LM & - & 1.00 & LM+Code & 0.59 & 1.00 \\
\hline
9 & LM & - & 1.00 & LM & - & 0.84 & LM & - & 0.93 \\
\hline
10 & LM & - & 1.00 & LM & - & 1.02 & LM & - & 1.22 \\
\hline
11 & LM & - & 1.00 & LM+Code & 0.66 & 0.77 & LM+Math & 0.66 & 1.00 \\
\hline
12 & LM & - & 0.91 & LM+Code & 0.60 & 1.00 & LM+Code & 0.60 & 1.13 \\
\hline
13 & LM+Code & 0.70 & 1.00 & LM+Math & 0.60 & 1.00 & \begin{tabular}{c}LM+Math\\+Code\end{tabular} & 0.60 & 1.11 \\
\hline
14 & LM+Math & 0.70 & 1.00 & LM+Math & 0.60 & 1.00 & LM & - & 1.00 \\
\hline
15 & LM & - & 1.00 & LM+Math & 0.70 & 1.00 & LM+Math & 0.66 & 1.00 \\
\hline
16 & Base & - & 1.00 & LM+Math & 0.60 & 1.00 & LM+Math & 0.60 & 1.00 \\
\hline
17 & LM & - & 1.00 & LM & - & 1.00 & LM & - & 1.00 \\
\hline
18 & LM & - & 1.00 & \multicolumn{3}{c|}{REMOVED} & \multicolumn{3}{c}{REMOVED} \\
\hline
19 & LM+Code & 0.70 & 1.00 & LM+Code & 0.60 & 1.00 & LM+Code & 0.60 & 1.01 \\
\hline
20 & LM+Code & 0.70 & 1.00 & LM & - & 1.00 & \multicolumn{3}{c}{REMOVED} \\
\hline
21 & LM & - & 1.00 & Base & - & 1.07 & Base & - & 1.07 \\
\hline
22 & LM & - & 1.00 & Math & - & 1.00 & LM+Math & 0.60 & 1.09 \\
\hline
23 & LM & - & 1.00 & \multicolumn{3}{c|}{REMOVED} & \multicolumn{3}{c}{REMOVED} \\
\hline
24 & LM & - & 1.00 & Base & - & 1.01 & Base & - & 1.01 \\
\hline
25 & \multicolumn{3}{c|}{REMOVED} & \multicolumn{3}{c|}{REMOVED} & \multicolumn{3}{c}{REMOVED} \\
\hline
26 & \multicolumn{3}{c|}{REMOVED} & LM & - & 1.04 & LM & - & 1.04 \\
\hline
27 & \multicolumn{3}{c|}{REMOVED} & \multicolumn{3}{c|}{REMOVED} & \multicolumn{3}{c}{REMOVED} \\
\hline
28 & \multicolumn{3}{c|}{REMOVED} & \multicolumn{3}{c|}{REMOVED} & \multicolumn{3}{c}{REMOVED} \\
\hline
29 & \multicolumn{3}{c|}{REMOVED} & \multicolumn{3}{c|}{REMOVED} & \multicolumn{3}{c}{REMOVED} \\
\hline
30 & \multicolumn{3}{c|}{REMOVED} & Base & - & 1.00 & Base & - & 1.00 \\
\hline
31 & \multicolumn{3}{c|}{REMOVED} & \multicolumn{3}{c|}{REMOVED} & \multicolumn{3}{c}{REMOVED} \\
\hline
32 & \multicolumn{3}{c|}{REMOVED} & \multicolumn{3}{c|}{REMOVED} & LM & - & 1.00 \\
\hline
33 & \multicolumn{3}{c|}{REMOVED} & \multicolumn{3}{c|}{REMOVED} & \multicolumn{3}{c}{REMOVED} \\
\hline
34 & LM & - & 1.00 & Base & - & 1.00 & Code & - & 1.00 \\
\hline
35 & Base & - & 1.00 & LM & - & 1.13 & LM & - & 1.28 \\
\hline
36 & LM & - & 1.00 & \multicolumn{3}{c|}{REMOVED} & \multicolumn{3}{c}{REMOVED} \\
\hline
37 & LM & - & 1.00 & LM & - & 1.00 & LM & - & 1.00 \\
\hline
38 & LM & - & 0.75 & LM & - & 1.00 & Math & - & 1.00 \\
\hline
39 & \multicolumn{3}{c|}{REMOVED} & Math & - & 1.00 & Math & - & 1.00 \\
\hline
\end{tabular}
}
\label{tab:arch13}
\end{table}

\begin{table}[H]
\centering
\caption{Architecture Parameters of pruned 7B models}
\resizebox{\textwidth}{!}{%
\begin{tabular}{c|c|c|c|c|c|c|c|c|c}
\hline
\rowcolor{lightgray}
\multirow{1}{*}{\textbf{Layer}} & \multicolumn{3}{c|}{\textbf{Model-1}} & \multicolumn{3}{c|}{\textbf{Model-2}} & \multicolumn{3}{c}{\textbf{Model-3}} \\
\cline{2-10}
\rowcolor{lightgray}
 & \textbf{Type} & \begin{tabular}{c}\textbf{Merge}\\\textbf{Factor}\end{tabular} & \begin{tabular}{c}\textbf{Output}\\\textbf{Scale}\end{tabular} & \textbf{Type} & \begin{tabular}{c}\textbf{Merge}\\\textbf{Factor}\end{tabular} & \begin{tabular}{c}\textbf{Output}\\\textbf{Scale}\end{tabular} & \textbf{Type} & \begin{tabular}{c}\textbf{Merge}\\\textbf{Factor}\end{tabular} & \begin{tabular}{c}\textbf{Output}\\\textbf{Scale}\end{tabular} \\
\hline
0 & LM & - & 1.00 & Math+Code & 0.48 & 1.00 & LM+Math & 0.48 & 0.92 \\
\hline
1 & LM+Math+Code & 0.50 & 1.00 & LM & - & 1.00 & LM & - & 1.00 \\
\hline
2 & LM & - & 1.03 & LM+Code & 0.52 & 1.06 & LM & - & 1.03 \\
\hline
3 & LM & - & 1.00 & Base & - & 0.98 & Math & - & 1.05 \\
\hline
4 & LM & - & 1.04 & LM & - & 1.11 & LM & - & 1.11 \\
\hline
5 & LM+Code & 0.59 & 1.08 & LM+Math & 0.38 & 1.12 & LM & - & 1.13 \\
\hline
6 & Code & - & 1.19 & Math & - & 1.25 & Code & - & 1.11 \\
\hline
7 & Code & - & 0.88 & LM+Code & 0.50 & 0.77 & LM+Code & 0.50 & 0.77 \\
\hline
8 & LM & - & 1.28 & LM & - & 1.34 & LM & - & 1.19 \\
\hline
9 & LM & - & 0.86 & LM & - & 0.93 & LM+Code & 0.51 & 0.56 \\
\hline
10 & Base & - & 1.00 & LM & - & 1.00 & LM & - & 1.00 \\
\hline
11 & LM+Math & 0.50 & 1.00 & Math & - & 1.02 & LM & - & 1.05 \\
\hline
12 & LM & - & 1.00 & LM+Math & 0.41 & 0.99 & LM+Math & 0.41 & 1.00 \\
\hline
13 & Math & - & 1.00 & LM+Math & 0.50 & 1.20 & LM+Math & 0.58 & 1.20 \\
\hline
14 & LM+Math & 0.60 & 1.00 & LM & - & 1.00 & LM+Math & 0.54 & 1.00 \\
\hline
15 & LM & - & 1.18 & Code & - & 0.97 & Code & - & 1.05 \\
\hline
16 & LM+Math & 0.50 & 1.00 & LM+Math & 0.50 & 1.00 & LM+Math & 0.45 & 1.00 \\
\hline
17 & LM+Math+Code & 0.50 & 1.00 & Code & - & 1.00 & Math+Code & 0.50 & 1.00 \\
\hline
18 & Math+Code & 0.50 & 1.00 & Base & - & 1.00 & Base & - & 1.01 \\
\hline
19 & \multicolumn{3}{c|}{REMOVED} & \multicolumn{3}{c|}{REMOVED} & \multicolumn{3}{c}{REMOVED} \\
\hline
20 & \multicolumn{3}{c|}{REMOVED} & \multicolumn{3}{c|}{REMOVED} & \multicolumn{3}{c}{REMOVED} \\
\hline
21 & LM & - & 1.00 & \multicolumn{3}{c|}{REMOVED} & LM & - & 1.00 \\
\hline
22 & \multicolumn{3}{c|}{REMOVED} & \multicolumn{3}{c|}{REMOVED} & \multicolumn{3}{c}{REMOVED} \\
\hline
23 & \multicolumn{3}{c|}{REMOVED} & \multicolumn{3}{c|}{REMOVED} & \multicolumn{3}{c}{REMOVED} \\
\hline
24 & \multicolumn{3}{c|}{REMOVED} & LM & - & 1.00 & \multicolumn{3}{c}{REMOVED} \\
\hline
25 & \multicolumn{3}{c|}{REMOVED} & \multicolumn{3}{c|}{REMOVED} & \multicolumn{3}{c}{REMOVED} \\
\hline
26 & \multicolumn{3}{c|}{REMOVED} & \multicolumn{3}{c|}{REMOVED} & \multicolumn{3}{c}{REMOVED} \\
\hline
27 & LM & - & 1.00 & Base & - & 0.99 & LM & - & 0.99 \\
\hline
28 & \multicolumn{3}{c|}{REMOVED} & LM & - & 1.00 & \multicolumn{3}{c}{REMOVED} \\
\hline
29 & LM+Code & 0.50 & 1.00 & LM & - & 1.00 & LM+Code & 0.50 & 1.00 \\
\hline
30 & \multicolumn{3}{c|}{REMOVED} & \multicolumn{3}{c|}{REMOVED} & \multicolumn{3}{c}{REMOVED} \\
\hline
31 & LM+Math & 0.50 & 1.00 & \multicolumn{3}{c|}{REMOVED} & LM+Math & 0.50 & 1.00 \\
\hline
\end{tabular}%
}
\label{tab:arch7}
\end{table}

\end{document}